\documentclass{article}

 \usepackage[eandd, final]{neurips_2026}

\usepackage[utf8]{inputenc} 
\usepackage[T1]{fontenc}    
\usepackage{hyperref}       
\usepackage{url}            
\usepackage{booktabs}       
\usepackage{amsfonts}       
\usepackage{nicefrac}       
\usepackage{microtype}      
\usepackage{xcolor}         
\usepackage{multirow}
\usepackage{wrapfig}
\usepackage{subcaption}
\usepackage{mathtools}
\usepackage{pifont}
\usepackage{natbib}
\usepackage[dvipsnames]{xcolor} 
\usepackage{graphicx}

\title{MoT3DVG: A Benchmark for Outdoor 3D Visual Grounding with Motion-Aware Descriptions and Temporal Cues}

\author{
  Shijia Zhao$^{ab}$\thanks{These authors contributed equally to this work.},
  Xusheng Guo$^{ab}$\footnotemark[1],
  Qiming Xia$^{ac}$\thanks{Corresponding authors},
  Xun Huang$^{abd}$,
  \\
  \textbf{Kezheng Xiong}$^{ab}$,
  \textbf{Zhihong Liu}$^{ab}$,
  \textbf{Chenglu Wen}$^{ab}$\footnotemark[2]\\
  $^{a}$Fujian Key Laboratory of Urban Intelligent Sensing and Computing, \\
  Xiamen University, Xiamen, China. \\
  $^{b}$Key Laboratory of Multimedia Trusted Perception and Efficient Computing, \\
  Ministry of Education of China, Xiamen University, Xiamen, China. \\
  $^{c}$The Hong Kong Polytechnic University, Hong Kong SAR, China. \\
  $^{d}$Zhongguancun Academy, China. \\
  \texttt{\{zhaoshijia4567,xusheng,huangxun,xiongkezheng,liuzh\}@stu.xmu.edu.cn}\\
  \texttt{qiminxia@polyu.edu.hk}\\
  \texttt{clwen@xmu.edu.cn}\\
}

\begin{document}

\maketitle

\begin{abstract}
3D visual grounding (3DVG) localizes language-referred objects in 3D scenes, with broad applicability in both indoor and outdoor environments. However, existing 3DVG research focuses on static indoor objects, limiting its use in dynamic outdoor autonomous driving.
Motion descriptions, which capture the temporal dynamics of moving objects, offer a promising way to address this challenge. We therefore introduce \emph{\textbf{MoT3DVG}}, a large-scale dataset for dynamic-aware outdoor 3DVG with temporally evolving motion descriptions. It contains 850 scenarios with 31,128 frames from nuScenes dataset, and provides 144,568 language prompts with motion-aware descriptions for dynamic objects across time.
We further propose \textbf{DynaVG}, a novel framework that effectively leverages temporal cues for outdoor 3DVG.
 In addition to the standard modality-specific encoders, DynaVG introduces a Short-Long Integrated Dynamic Encoding (SLIDE) module before the language-point cloud alignment.
SLIDE models both local motion cues and global temporal context via short-term window encoding and long-term window shifting.
Extensive experiments demonstrate its state-of-the-art performance on \emph{\textbf{MoT3DVG}}, validating the effectiveness of motion-aware encoding. Our work reveals open challenges and promising directions for future research in outdoor 3DVG.
\end{abstract}
\section{Introduction}
Given a natural language description, 3D visual grounding (3DVG) aims to localize the referred object within a 3D scene,
 and thus constitutes a fundamental prerequisite for downstream tasks such as embodied navigation, 3D visual question answering, and 3D scene graph generation \cite{3dvqa, drivemlm, lmdrive, mono3dvlt, msgnav, huang2026beyond}.

\setlength{\tabcolsep}{6pt}

\begin{table*}[htbp]
\caption{
Comparison of outdoor language-based datasets, where "C" and "L" mean Camera and LiDAR; "M.V.", "Mot.", "T. Mot", "Expr.", and "Obj." are short for multi-view, motion, temporal motion, expressions, and objects, respectively. Our dataset focuses specifically on language descriptions of the continuous dynamic behaviors exhibited by outdoor objects.
}
\centering
\small
\setlength{\tabcolsep}{1.5mm}
\begin{tabular}{c|c|cccc|cccc}
\toprule
Dataset     & Sensor & 3D                         & M.V.                 & Mot.                     & T. Mot.             & \multicolumn{1}{c}{Expr.} & \multicolumn{1}{c}{Frames} & \multicolumn{1}{c}{Obj.} & 
\multicolumn{1}{c}{Scenes}
\\ \midrule \midrule
Talk2Car \cite{talk2car}   & C      & \textcolor{red}{\ding{55}} & \textcolor{red}{\ding{55}} & \textcolor{Green}{\ding{51}} & \textcolor{red}{\ding{55}} & 11.9K                     & 9,217                      & 10.5K  & -                  \\
Refer-KITTI \cite{refer-kitti} & C      & \textcolor{red}{\ding{55}} & \textcolor{red}{\ding{55}} & \textcolor{Green}{\ding{51}} & \textcolor{red}{\ding{55}} & 0.8K                      & 6,650                      & -  & 18                     \\
Mono3DRefer \cite{mono3dvg} & C      & \textcolor{Green}{\ding{51}} & \textcolor{red}{\ding{55}} & \textcolor{Green}{\ding{51}} & \textcolor{red}{\ding{55}} & 41K                       & 2,025                      & 8.2K  & -                    \\
Talk2Car-3D \cite{talk2lidar} & L + C    & \textcolor{Green}{\ding{51}} & \textcolor{red}{\ding{55}} & \textcolor{Green}{\ding{51}} & \textcolor{red}{\ding{55}} & 8.4K                      & 5,534                      & - & -                   \\
TOD$^3$Cap \cite{tod3cap}  & L + C    & \textcolor{Green}{\ding{51}} & \textcolor{Green}{\ding{51}} & \textcolor{Green}{\ding{51}} & \textcolor{red}{\ding{55}} & 2.3M                      & 34,149                     & - & 850                        \\
NuPrompt \cite{nuprompt}   & L + C    & \textcolor{Green}{\ding{51}} & \textcolor{Green}{\ding{51}} & \textcolor{Green}{\ding{51}} & \textcolor{red}{\ding{55}} & 35K                       & 34,149                     & - & 850                        \\
NuGrounding \cite{nugrounding} & L + C    & \textcolor{Green}{\ding{51}} & \textcolor{Green}{\ding{51}} & \textcolor{Green}{\ding{51}} & \textcolor{red}{\ding{55}} & 2.2M                      & 34,149                     & - & 850                       \\
3EED \cite{3eed}       & L + C    & \textcolor{Green}{\ding{51}} & \textcolor{red}{\ding{55}} & \textcolor{Green}{\ding{51}} & \textcolor{red}{\ding{55}} & 22K                       & 20,367                     & 128K  & -                    \\ \midrule
\textbf{\emph{MoT3DVG}} (\textbf{Ours})     & L + C    & \textcolor{Green}{\ding{51}} & \textcolor{Green}{\ding{51}} & \textcolor{Green}{\ding{51}} & \textcolor{Green}{\ding{51}} & 144K                      & 31,128                     & 985K   & 850                   \\ \bottomrule
\end{tabular}

\label{tab:benchmark_compare}
\end{table*}
\begin{wrapfigure}{R}{0.55\textwidth}
\vspace{-12.5pt}
    \includegraphics[width=0.55\textwidth]{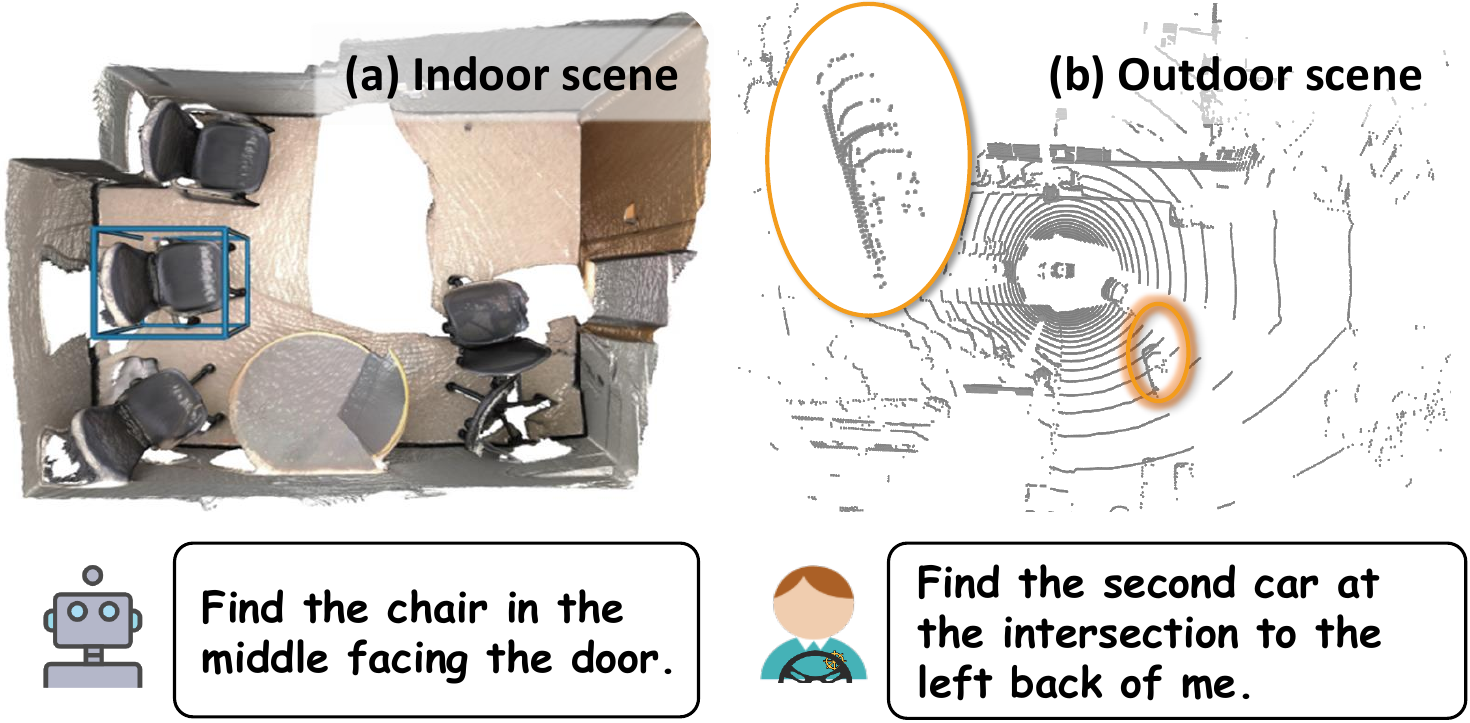}
  
   \caption{Objects in indoor and outdoor scenarios.
   (a) In most cases, spatial relationships among indoor objects remain static in short time intervals, whereas (b) objects in outdoor scenes exhibit non-fixed spatial relationships due to their dynamic behaviors.
}
   \label{fig:motivation}
\vspace{-1.0em}
\end{wrapfigure}

Most existing 3DVG works focus on indoor scenes and perform object localization in 3D environments via language–visual feature alignment \cite{bdetr, mvt, eda, zsvg3d, seeground}.
These approaches largely focus on static spatial relationships, such as those among objects and between objects and the surrounding scene.
As illustrated in Fig.~\ref{fig:motivation} (a), indoor objects typically exhibit negligible motion over short time windows, making static relations often sufficient for 3DVG models.
However, due to the sparsity and non-uniformity of outdoor point clouds, indoor methods are difficult to transfer directly to outdoor scenes.

To advance outdoor 3DVG, recent work introduces benchmarks by augmenting existing large-scale autonomous driving datasets with language descriptions for object bounding boxes \cite{talk2car, lidar-llm}. 
Furthermore, 3EED \cite{3eed} introduces multi-scale and scale-aware fusion strategies to mitigate the sparsity of the outdoor point cloud.
However, these methods remain limited to static object attributes and overlook the dynamic nature of outdoor scenarios.
As illustrated in Fig.~\ref{fig:motivation} (b), outdoor objects can undergo substantial motion even within short temporal intervals.
Such dynamics break the assumptions of (quasi-)static relationships. 
For example, a car described as “the car to the left of the building” in one frame may become “the car in front of the utility pole” in the next.
Motion attributes can effectively eliminate such semantic ambiguity arising from spatial uncertainty \cite{motionbench}.
Based on this, we conduct this work with two unique challenges for outdoor 3DVG:
\emph{i)} \textbf{Missing temporal motion descriptions.} Existing benchmarks are predominantly snapshot-based and often describe motion using coarse predicates (e.g., \emph{a moving car}), thereby preventing models from exploiting the fine-grained dynamics of outdoor objects.
\emph{ii)} \textbf{Observation-semantic misalignment.} Single-frame point cloud discards temporal cues in encodings, leading to a mismatch between the observation and motion-aware language description.

To facilitate 3DVG research in dynamic outdoor environments, we introduce \emph{\textbf{MoT3DVG}}, a large-scale dataset with language descriptions of continuous motion behaviors of dynamic objects.
Built upon nuScenes \cite{nuscenes}, \emph{\textbf{MoT3DVG}} contains over 985,000 objects and 144,568 descriptions with 31,128 frames across 850 scenes, capturing how motion states evolve over time. 
We present a comparison of our benchmark against existing outdoor language-based benchmarks in Tab. \ref{tab:benchmark_compare}.

We also propose \textbf{DynaVG}, a concise framework to achieve \textbf{Dyna}mic-aware learning for outdoor 3D\textbf{VG}.
As shown in Fig. \ref{fig:framework}, DynaVG consists of modality-specific encoders that extract point cloud and language features, respectively; a cross-modality encoder that performs point cloud and language feature interaction; a decoder; and a grounding head that outputs the referred 3D box.
To inject temporal motion cues into the current-frame representation, we introduce a Short-Long Integrated Dynamic Encoding (\textbf{SLIDE}) module after point cloud encoding.
SLIDE contains two components: short-term window encoding (STWE) and long-term window shifting (LTWS).
Specifically, STWE groups the historical features into chronologically ordered, motion-aware windows. Each window aggregates densely-sampled frames, enabling feature interactions to capture short-range motion patterns. 
LTWS then shifts these windows along the temporal axis to enlarge the temporal receptive field. The resulting long-range temporal cues are fused into the current-frame features via cross-attention, enabling motion-aware grounding in dynamic outdoor scenes.
Extensive comparisons with state-of-the-art indoor and outdoor 3DVG methods show that DynaVG achieves robust visual grounding in dynamic outdoor environments.
Our contributions are summarized as follows:
\begin{itemize}
    \item We introduce \emph{\textbf{MoT3DVG}}, the first large-scale outdoor 3DVG dataset with fine-grained motion-aware language descriptions of moving objects.
    Built upon nuScenes, \emph{\textbf{MoT3DVG}} covers over 985,000 objects and 144,568 motion-aware language descriptions across time for dynamic objects.
    \item We propose \textbf{DynaVG}, a dynamic-aware outdoor 3DVG framework. Its core component, the Short-Long Integrated Dynamic Encoding (\textbf{SLIDE}) module, injects historical temporal cues into current-frame point cloud features to achieve semantic alignment between language and point clouds.
    \item We benchmark state-of-the-art methods under multiple settings, demonstrating the importance of temporal cues and providing insights for future outdoor 3DVG research.
\end{itemize}
\section{Related Work}

\textbf{3D Object Detection.} 
3D object detection aims to identify and localize objects from predefined categories within a perceived 3D scene \cite{hinted, l4dr, winmamba}. Classical 3D detectors embrace one-stage \cite{voxelnet, pointpillars, centerpoint, dsvt}, two-stage \cite{pv-rcnn, voxel-rcnn, sparse, ted, owl}, or multi-stage designs \cite{cai20223d, casa, 3d-hanet}, using either point cloud or camera modalities to regress object locations and bounding boxes with deep networks. 
To compensate for the limitations of a single modality, recent studies have explored multi-modal fusion approaches to enhance the robustness of 3D detectors \cite{virtual-conv, bevfusion, graphbev, adaco, sp3d}.
Although these methods have demonstrated superior performance in the category-level detection task, the absence of language guidance prevents them from precisely localizing specific objects.

\textbf{Visual Grounding in 3D Environments.}
Compared with category-level object detection, 3D visual grounding uses natural language prompts to precisely localize a specific referred object in a 3D scene.
Early efforts predominantly focused on indoor scenes.
ScanRefer \cite{scanrefer} established the first large-scale indoor dataset that aligns 3D environment semantics with free-form descriptions by annotating objects in ScanNet.
ScanRefer, along with the subsequently proposed Sr3D and Nr3D \cite{referit3d}, has become widely-used benchmarks suite for indoor 3DVG \cite{sat, 3dvg-transformer, bdetr, 3d-vista, eda, zsvg3d, augrefer, seeground, liba}.

Recent efforts have begun to extend 3DVG to outdoor scenarios \cite{kitti360pose, wildrefer, mono3dvg}.
By supplementing nuScenes \cite{nuscenes} with language annotations, Talk2Car \cite{talk2car} established the early 3DVG dataset for autonomous driving scenes.
Based on this dataset, Talk2LiDAR \cite{talk2lidar} explores outdoor grounding by leveraging multi-modal fusion strategies and the CLIP model.
Talk2Radar \cite{talk2radar} investigates robust text–point cloud alignment and fusion based on 4D radar.
LiDAR-LLM \cite{lidar-llm} endows large models with 3D scene understanding through large-model fine-tuning and a three-stage training strategy.
Furthermore, 3EED \cite{3eed} extends 3DVG to a multi-platform benchmark via cross-platform data alignment.
However, these works typically rely on snapshot observation and thus overlook the continuous motions of outdoor objects.

\textbf{Language Prompt Motion Perception in Outdoor 3D Environments.} 
Recently, some outdoor language prompt-based works have begun to incorporate motion information of objects. TOD$^3$Cap \cite{tod3cap} treats movement as an object attribute and generates a dense caption for each object in the scene.
Similarly, 3EED \cite{3eed} embeds motion cues into the language prompt.
Talk2Radar \cite{talk2radar} emphasizes the importance of object motion for outdoor navigation and accordingly introduces the 4D radar modality to capture object motion.
NuPrompt \cite{nuprompt} utilizes a language prompt to forecast the trajectory of objects across multi-view images.
Although these works have recognized the dynamic characteristics of outdoor scenes, they often treat motion as a coarse attribute rather than the continuous temporal evolution.

\begin{figure*}[t]
  \centering
   \includegraphics[width=0.99\linewidth]{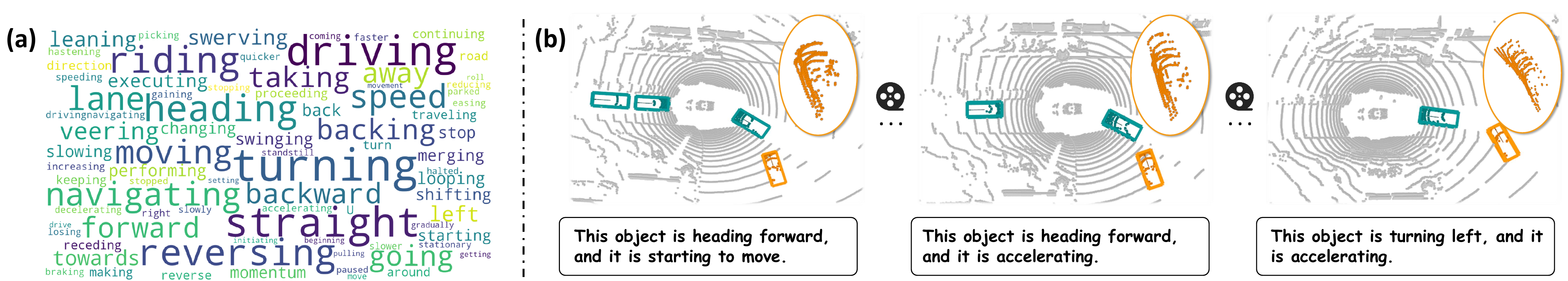}
   \caption{
   The visualization of \textbf{\emph{MoT3DVG}}. (a) present the word cloud of the dataset, which encompasses diverse motion states of dynamic objects; (b) present examples of temporal motion description, where orange denotes the language-referred object, while green indicates non-referred objects.
   }
   \label{fig:word_cloud_and_vis}
\end{figure*}

\section{\emph{\textbf{MoT3DVG}} Dataset}

In outdoor scenes, existing 3DVG datasets typically focus solely on static attributes such as object category, appearance, and instantaneous spatial position.
Although these works account for coarse motion descriptions, they overlook the temporal evolution of object motions.
To this end, we present the \emph{\textbf{MoT3DVG}} dataset, which enriches referring expressions with motion-aware descriptions of dynamic objects for language–point cloud semantic matching in outdoor 3DVG.
Built on the widely used nuScenes dataset \cite{nuscenes}, \emph{\textbf{MoT3DVG}} provides motion-aware language descriptions for dynamic objects at specific timestamps.
The data collection pipeline is detailed in Sec. \ref{sec: data_collection} and dataset statistics are provided in Sec. \ref{sec: data_statistics}.

\subsection{Data Collection}\label{sec: data_collection}
\textbf{Collection Principle.}
When humans describe an object, they usually mention its category, color, appearance, and spatial position (e.g., “\emph{This is a black trash can. It is under a bathroom counter.}”). 
However, in outdoor scenes, numerous objects can share similar attributes, making it easy to confuse them when relying solely on the aforementioned static properties.
Due to the inherent dynamics of outdoor scenes, the motion behaviors of objects provide an additional cue that disambiguates different objects.
This motivates us to extend object descriptions with motion information.

Recently, TOD$^3$Cap \cite{tod3cap} leverages a large multi-modal model and human verification to provide high-quality caption annotations for objects in nuScenes.
However, the motion status of objects is only vaguely covered; it is labeled as either \emph{moving} or \emph{not moving}, leaving their motion information unexplored.
In \emph{\textbf{MoT3DVG}}, we decompose a detailed temporal motion description into \textbf{directional} and \textbf{speed} components.
Specifically, the directional description characterizes object behavior along the motion direction, while the speed description depicts changes in object velocity.
We then combine the two components by instantiating a predefined motion template $\mathcal{T}_m$:
\begin{equation}
\mathcal{C}_m = \mathcal{T}_m(\mathcal{C}_d, \mathcal{C}_s),
\label{eq:motion_template}
\end{equation}
where $\mathcal{T}_m(\cdot)$ maps directional descriptions $\mathcal{C}_d$ and speed description $\mathcal{C}_s$ to a final motion description $\mathcal{C}_m$.

\textbf{Data Annotation.}
We adopt a simple three-step strategy to obtain the motion information of objects in outdoor 3D environments.
1) \emph{Dynamic-static object filtering}: For each instance, we first compute its centroid and estimate its velocity. If its average displacement and average speed over $T$ timestamps are below the preset thresholds, we classify it as static and filter it out.
2) \emph{Temporal frame stitching}: For each scene in nuScenes, we transform the continuous point cloud sequences into the world coordinate system and concatenate them. 
We then transform each object's trajectory back to the first frame coordinate system using the inverse extrinsic of the first frame, so that motion can be consistently measured.
3) \emph{Motion annotation}: 
Equipped with temporal information, we project the trajectory of each object onto the BEV plane. 
Human annotators then label the directional and speed components at each timestamp, with cross-checking performed among them. More details can be found in the appendix \ref{sec: supp_annotation}.

\subsection{Data statistics}\label{sec: data_statistics}
We annotate the dynamic objects in the nuScenes~\cite{nuscenes} with 31,128 frames across 850 scene sequences.
Built upon nuScenes, \textbf{\emph{MoT3DVG}} includes over 985,000 objects and 144,568 temporally grounded motion-aware descriptions, each of which refers to a target object at a specific timestamp.
Adhering to the official split of the nuScenes dataset, we divided \textbf{\emph{MoT3DVG}} into 117,072 descriptions for training across 700 scenes and 27,496 descriptions for validation across 150 scenes.
In each point cloud frame, we designate one described object as the language referent and the others as non-referents. 
Fig. \ref{fig:word_cloud_and_vis} shows the visualization in our dataset, including the word cloud and examples of temporal motion descriptions. During annotation, we provide diverse motion description templates to mitigate repetitive phrasing. More details can be found in the appendix \ref{sec: supp_data_statistic}.
\section{DynaVG Network}
\begin{figure*}[t]
  \centering
   \includegraphics[width=0.99\linewidth]{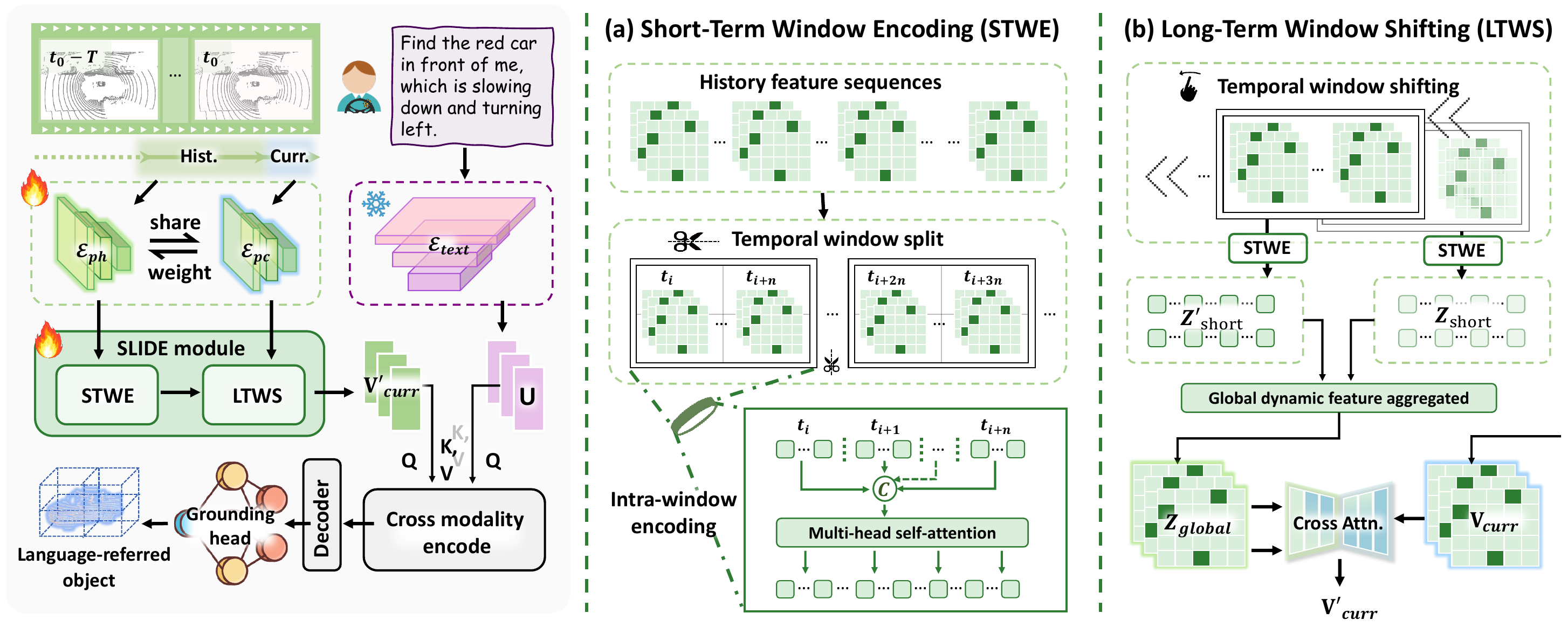}
   \caption{The overview of proposed \textbf{DynaVG}. It takes a temporal sequence of point cloud frames and a language description as input. Leveraging the proposed SLIDE module, DynaVG encodes historical temporal cues into current frame point features, fuses them with the language embedding, and decodes the language-referred object.
   }
   \label{fig:framework}
\end{figure*}

\subsection{Preliminaries}
\textbf{Problem Definition.}
3D visual grounding aims to accurately localize a specific object in a 3D environment given a language reference.
Specifically, the model is trained on strictly aligned language–point cloud pairs. Given a language caption $c\in \mathcal{C}$ and a 3D scene $s\in \mathcal{S}$, it predicts the bounding box $b\in \mathcal{B}$ of the language-referred object, including its localization and size:
\begin{equation}
\mathcal{G}_\theta : \mathcal{C} \times \mathcal{S} \to \mathcal{B},
\label{eq:grounding_define}
\end{equation}
where $\mathcal{G}_\theta$ denotes the grounding model.

\textbf{Overview.}
In this paper, we propose a concise framework, termed \textbf{DynaVG}, to utilize temporal motion cues to boost outdoor 3DVG. The pipeline of this framework is shown in Fig. \ref{fig:framework}.
Most existing 3DVG models process a single-frame point cloud alongside the grounding utterance. 
With motion-bearing phrases, such as “\emph{turning left}", the model can hardly infer the motion state of the object from a solitary snapshot, which leads to semantic misalignment between point clouds and language prompts.
The core motivation of DynaVG is to enrich the current frame point cloud features with temporal cues from historical frames, better aligning them with motion-aware language queries.

\subsection{Modality-specific Encoder}
\textbf{Point Cloud Encoder.}
Indoor scenarios typically exhibit uniformly dense point clouds and limited object-scale variation. In stark contrast, outdoor environments are open and yield sparse point clouds with extreme object-scale variations.
Built on PointNet++ \cite{pointnet++}, 3EED \cite{3eed} introduces multi-scale sampling and scale-aware fusion to address these challenges.
Using this enhanced PointNet++ model, we encode the input point cloud into visual embeddings $\mathbf{V}\in\mathbb{R}^{n_v\times c_v}$, where $n_v$ and $c_v$ denote the number of visual tokens and their dimensions, respectively.

To capture historical object motion, we anchor the sequence at the current-frame $t_0$ and input a point cloud sequence $\mathcal{P} = \lbrace \mathcal{P}_{hist}, \mathcal{P}_{curr} \rbrace$ of length $T+1$, where $\mathcal{P}_{hist} = \lbrace p_{t_0-T},p_{t_0-(T-1)},...,p_{t_0-1} \rbrace$ denotes the historical point cloud sequence, and $\mathcal{P}_{curr}=p_{t_0}$ is the current frame.
Specifically, we first align 
$\mathcal{P}_{hist}$ to the coordinate frame of $\mathcal{P}_{curr}$ according to ego-motion compensation. 
After that, using shared-weight point cloud encoders $\mathcal{E}_{ph}$ and $\mathcal{E}_{pc}$, we encode the $\mathcal{P}_{hist}$ and $\mathcal{P}_{curr}$ into $\mathbf{V}_{hist}\in\mathbb{R}^{T\times n_v\times c_v}$ and $\mathbf{V}_{curr}\in\mathbb{R}^{n_v\times c_v}$, respectively:
\begin{equation}
\mathbf{V}_{hist}=\mathcal{E}_{ph}(\mathcal{P}_{hist}),\;\mathbf{V}_{curr}=\mathcal{E}_{pc}(\mathcal{P}_{curr}).
\label{eq:pc_encode}
\end{equation}
This weight sharing enforces cross-frame temporal consistency in feature extraction.

\textbf{Language Encoder.}
Following \cite{bdetr, 3eed}, we encode the input utterance using a pre-trained natural language processing (NLP) model, yielding language embeddings $\mathbf{U}\in\mathbb{R}^{n_u\times c_l}$, where $n_u$ and $c_l$ denote the token sequence length and embedding dimension, respectively.
A linear projection then maps $\mathbf{U}$ to match the dimensionality of the point cloud features.

\subsection{Short-Long Integrated Dynamic Encoding}
Inspired by the local-global spatio-temporal modeling strategy in VST \cite{video-swint}, which fuses the entire sequential spatio-temporal features, we introduce a Short-Long Integrated Dynamic Encoding (\textbf{SLIDE}) module to embed the implicit motion information in $\mathbf{V}_{hist}$ into $\mathbf{V}_{curr}$.
Distinct from VST, our SLIDE concentrates on utilizing historical frame information as supplementary temporal cues to implicitly align motion-aware language prompts with current frame features.
Driven by this purpose, SLIDE is constructed with two core strategies: \textbf{short-term window encoding} and \textbf{long-term window shifting}, as illustrated in Fig. \ref{fig:framework}.

\textbf{Short-Term Window Encoding.}
Typically, object motion in a point cloud sequence exhibits bounded displacement over short time intervals $\Delta t$:
\begin{equation}
||l_o(t+\Delta t)-l_o(t)||_2 \leq v_{max}\cdot\Delta t,
\label{eq:short_motion_bound}
\end{equation}
where $l_o(t)$ represents the location of object $o$ at timestamp $t$, and $v_{max}$ is the maximum velocity of $o$.
Motivated by this property, modeling motion over short time windows effectively preserves salient motion cues of the object.
Specifically, we partition $\mathbf{V}_{hist}$ along the temporal axis into $\mathcal{K}=\lfloor T/w\rfloor$ non-overlapping windows.
For notational clarity, we use non-negative time index $i\in\lbrace0,1,\ldots,T-1\rbrace$ to denote the $i$-th historical frame:
\begin{equation}
\mathcal{W}_k=\lbrace i\in\lbrace0,\ldots, T-1,\rbrace\;|\;kw\leq i < (k+1)w\rbrace,
\label{eq:window_split}
\end{equation}
where $w$ denotes the window size, specifically the number of frames per window, and $k\in\lbrace0,1,\ldots,\mathcal{K}-1\rbrace$ is the window index.
Within each short temporal window, we first construct a token sequence using the features of adjacent frames:
\begin{equation}
\mathbf{X}^k=\mathcal{F}(\mathbf{V}_{hist[kw:(k+1)w],:,:]})\in\mathbb{R}^{(w\times n_v)\times c_v},
\label{eq:window_cat}
\end{equation}
where $\mathcal{F}(\cdot)$ is the tensor flattening operation to form a token sequence. We then apply a multi-head self-attention module $\text{MHA}^{intra}(\cdot)$ to aggregate motion-aware features:
\begin{equation}
\mathbf{Z}_{k}=\text{MHA}^{intra}(\mathbf{X}^{k})\in\mathbb{R}^{(w\times n_v)\times c_v}.
\label{eq:mha}
\end{equation}
The final encoded representation $\mathbf{Z}_{short}=[\mathbf{Z}_0,\ldots,\mathbf{Z}_{\mathcal{K}-1}]$ for the entire history is obtained by concatenating across the windows.

\textbf{Long-Term Window Shifting.}
Although the above strategy effectively models interactions among adjacent frames within each window, interactions between adjacent frames across window boundaries are largely neglected due to the hard window partitioning.
We therefore introduce window shifting to capture long-term temporal cues through cross-window interactions.
Specifically, we first apply a temporal shift to the historical frame features along the time axis.
We define a circular temporal shift operator on the historical features $\mathbf{V}_{hist}$:
\begin{equation}
[\mathcal{R}^{\alpha}_t(\mathbf{V}_{hist})]_{i,:,:} = \mathbf{V}_{hist\,((i+\alpha)\bmod T),:,:},
\label{eq:window_rolling}
\end{equation}
\setlength{\tabcolsep}{6pt}

\begin{table*}[htbp]
\caption{
Benchmark results of state-of-the-art 3DVG methods on the \textbf{\emph{MoT3DVG}} dataset.
We report accuracy, which is evaluated by distances of 1.0 and 0.5m, and mIoU across the predicted grounding boxes for \emph{Unique} subset, \emph{Multiple} subset, and \emph{Overall}. All metrics are presented in percentage (\%).
}
\centering
\small
\setlength{\tabcolsep}{1mm}
\begin{tabular}{c|c|ccc|ccc|ccc}
\toprule
\multirow{2}{*}{Method} & \multirow{2}{*}{Setting}& \multicolumn{3}{c|}{Unique}       & \multicolumn{3}{c|}{Multiple} & \multicolumn{3}{c}{Overall}         \\  &
                  & @1.0  & @0.5 & mIoU & @1.0 & @0.5 & mIoU  & @1.0 & @0.5 & mIoU \\ \midrule \midrule 
BUTD-DETR \cite{bdetr}  & \multirow{3}{*}{Indoor}      & 17.00    & 8.75    & 7.30        & 26.10     & 15.42        & 9.25        & 44.27     & 28.87  & 15.38     \\
EDA \cite{eda}        &      & 36.34    &  15.54   & 13.08       & 52.43     & 36.57     & 16.95      & 50.74     & 35.80     & 16.49     \\ 
AugRefer \cite{augrefer}&      & 32.83    &  15.54   & 12.23       & 53.75     & 38.50     & 17.11      & 54.59     & 37.31     & 17.48     \\ 
\midrule
WildRefer \cite{wildrefer} & \multirow{3}{*}{Outdoor}       & 15.04      & 8.02    & 9.79    & 35.47     & 18.60      & 14.24        &33.40         & 17.65         & 14.07     \\
3EED \cite{3eed}  &           & 51.38    & \textbf{31.58}    & 16.73        & 58.45     & 47.96        & 18.23        & 67.22        & 54.94 & 21.16     \\
\textbf{DynaVG (Ours)} &   & \textbf{51.63}    & 29.32   & \textbf{20.21} & \textbf{71.21}    & \textbf{60.25}        & \textbf{22.13}        & \textbf{69.66}  & \textbf{57.34}  & \textbf{22.22}    \\ \bottomrule
\end{tabular}

\label{tab:sota_comp}
\end{table*}
where $i$ is the timestep index and $\alpha$ is the shift size. We then re-partition the shifted sequence into windows and repeat the intra-window encoding, enabling cross-window interactions.
We then fuse the newly aggregated motion-aware features $\mathbf{Z}'_{short}$ with the features from the previous stage $\mathbf{Z}_{short}$, similar to EMA, to obtain refined global dynamic features:
\begin{equation}
\mathbf{Z}_{global}=\beta \cdot\mathbf{Z}'_{short}+(1-\beta)\cdot\mathbf{Z}_{short},
\label{eq:ema}
\end{equation}
where $\beta\in(0,1)$ is a learnable scalar that controls the contribution of the shifted aggregation.
Finally, we embed the dynamic information into the current-frame features via a cross-attention module.

\subsection{Cross-modality Interaction}
\textbf{Cross-modality Encode.}
This stage constructs a shared semantic subspace via a bidirectional cross-attention module~\cite{bdetr, eda, 3eed}, enabling interaction between point cloud and language features.
As shown in Fig. \ref{fig:framework}, this module takes the aligned language and point cloud embeddings as keys and values, and uses each modality as queries to attend to and aggregate information from the other.

\textbf{Decoder.}
Following \cite{bdetr, 3eed}, we select $K_s$ seed points from the semantically aligned point cloud features to generate the initial query set.
These queries represent potential foreground objects and serve as coarse anchors for subsequent decoding.
In each decoder layer, this query set is iteratively refined
through interactions with both language and point cloud features in accordance with the attention mechanism.
Finally, the refined query set is fed into a prediction head to output the location and size of the object referred to by the language prompt.
\section{Experiments}

\subsection{Implementation Details}
Following \cite{bdetr, eda, 3eed}, we employ RoBERTa \cite{roberta} as the language encoder and keep its parameters frozen during training.
During point cloud encoding, we treat the historical-frame branch as an auxiliary pathway and detach it from the gradient computation, enhancing model stability while reducing computational cost.
In the partition of the motion-aware windows in the SLIDE module, we set $\mathcal{K}=2$ and $w=3$ to capture motion spanning approximately 2 to 3 seconds.
During window shifting, we set $\alpha=1$ and apply the shift once before re-partitioning, so that the windows are offset by one frame to enable cross-window interaction.
The learning rate is set to 0.001 for the point cloud encoder and 0.0001 for the others, and we train for 100 epochs with a batch size of 8 across four NVIDIA RTX 3090 GPUs.

\textbf{Evaluation Metrics.}
To evaluate grounding models on the \textbf{\emph{MoT3DVG}}, we follow the nuScenes protocol \cite{nuscenes} and report Top-1 accuracy, counting a success when the Euclidean distance between the predicted location and the ground truth is within a predefined threshold.
We evaluate at Acc$\text{@}m$, $m\in[0.5, 1.0]$ meters, and report the mean 3D Intersection over Union (mIoU) for overall quality.

\setlength{\tabcolsep}{6pt}

\begin{table*}[htbp]
\caption{
Comparison with state-of-the-art 3DVG method on \textbf{\emph{MoT3DVG}} across various distance ranges.
We partition the dataset into three distance intervals according to the valid range of the dataset to investigate the impact of varying distances on model performance.
}
\centering
\small
\setlength{\tabcolsep}{1mm}
\begin{tabular}{c|c|ccc|ccc|ccc}
\toprule
\multirow{2}{*}{Method} &
\multirow{2}{*}{Range} &
\multicolumn{3}{c|}{Unique} &
\multicolumn{3}{c|}{Multiple} &
\multicolumn{3}{c}{Overall}\\
&
&
\multicolumn{1}{c}{@1.0} & @0.5 & mIoU$\;$ &
@1.0 & @0.5 & mIoU$\;$ &
@1.0 & @0.5 & mIoU$\;$
\\ \midrule \midrule
3EED \cite{3eed}  & \multirow{3}{*}{0-20m}    & 57.08        & 43.38        & 17.88$\;$     & 64.64 & 56.36 & 18.73$\;$ & 73.23  & 64.55 & 21.46$\;$
\\
DynaVG (Ours) &         & 50.68        & 38.81     &  19.02$\;$ & 75.95  & 67.43 & 22.02$\;$  & 74.22  & 64.87 & 21.81$\;$     \\ 
\emph{Improvement} & & -6.40        & -4.57        & +1.14     & +11.31 & +11.07  & +3.29$\;$  & +0.99  & +0.32  & +0.35$\;$     \\
\midrule
3EED \cite{3eed}   & \multirow{3}{*}{20-40m} &  43.79       & 18.30        & 14.45$\;$     & 48.95 & 32.38  & 18.32$\;$  & 57.02 & 37.24 & 21.27$\;$      \\
DynaVG (Ours) &         & 52.29        & 21.57     & 19.35$\;$  & 59.71  & 43.71 & 22.25$\;$  & 59.93  & 40.32  & 22.51$\;$    \\ 
\emph{Improvement} & & +8.50        & +3.27        & +4.90$\;$     & +10.76 & +11.33  & +3.93$\;$  & +2.91  & +3.08 & +1.24$\;$     \\ 
\midrule
3EED \cite{3eed}   & \multirow{3}{*}{40-50m} & 10.00        & 6.67        & 5.40$\;$     & 15.28 & 5.56 & 8.81$\;$ & 12.12 & 6.06 & 11.41$\;$     \\
DynaVG (Ours) &         & 26.67        & 6.67     & 31.55$\;$ & 23.61 & 5.56 & 15.23$\;$  & 20.20  & 6.06 & 17.13$\;$       \\ 
\emph{Improvement} &         & +16.67        & +0.00     & +26.15$\;$  & +8.33 & +0.00  & +6.42 & +8.08  & +0.00  & +5.72$\;$     \\ 
\bottomrule
\end{tabular}

\label{tab:dist_compare}
\end{table*}
\begin{table}[t]
    \centering
    \begin{minipage}[t]{0.47\textwidth}
    \centering
    \caption{Comparison with SOTA 3DVG methods on 3EED vehicle dataset.
    Accuracy is reported at IoU 0.25 and 0.5 for fair comparison. 
    $\dag$ denotes our re-implemented results.}
    \vspace{0.1cm}
    \label{tab:3eed_compare}

    \resizebox{\textwidth}{!}{
    \begin{tabular}{c|ccc}
    \toprule
    Method  & @0.25 & @0.5 & mIoU  \\ 
    \midrule\midrule
    BUTD-DETR \cite{bdetr} & $52.38$ & $32.18$ & $-$ \\
    EDA \cite{eda}         & $53.54$ & $34.87$ & $-$ \\
    \midrule
    WildRefer \cite{wildrefer} & $50.27$ & $9.85$ & $-$ \\
    3EED \cite{3eed}           & $78.37$ & $45.72$ & $-$ \\
    3EED$^\dag$ \cite{3eed}    & $78.38$ & $45.07$ & $43.94$ \\
    \midrule
    \textbf{DynaVG (Ours)}     & $80.02$ & $46.97$ & $45.41$ \\
    \bottomrule
    \end{tabular}
    }
    \vspace{-1.0em}
    \end{minipage}
    \hfill
    \begin{minipage}[t]{0.5\textwidth}
    \centering
    \caption{Class-wise comparison on fine-grained vehicle categories
    "Cat." is the short form of category, "V.Car" and "V.Bus" denote fine-grained vehicle categories.}
    \vspace{0.12cm}
    \label{tab:class_wise_compare}

    \resizebox{\textwidth}{!}{
    \begin{tabular}{c|c|ccc}
    \toprule
    Method & Cat. & @1.0 & @0.5 & mIoU \\
    \midrule\midrule
    3EED \cite{3eed}  & \multirow{3}{*}{V.Car} & $64.28$ & $55.79$ & $19.46$ \\
    DynaVG (Ours)     &                        & $77.74$ & $69.85$ & $23.46$ \\
    \emph{Improvement}&                        & $+13.46$& $+14.06$& $+4.00$ \\
    \midrule
    3EED \cite{3eed}  & \multirow{3}{*}{V.Bus} & $43.64$ & $16.84$ & $19.08$ \\
    DynaVG (Ours)     &                        & $46.05$ & $18.56$ & $21.27$ \\
    \emph{Improvement}&                        & $+2.41$ & $+1.72$ & $+2.19$ \\
    \bottomrule
    \end{tabular}
    }
    \vspace{-1.0em}
    \end{minipage}
\end{table}

\textbf{Baselines.}
We adopt representative indoor and outdoor methods for benchmarking.
For indoor methods, we select BUTD-DETR~\cite{bdetr} and EDA~\cite{eda}, following the setup of 3EED~\cite{3eed}.
We primarily focus on the outdoor setting method, where we adapt the state-of-the-art 3EED \cite{3eed}.
It introduces multi-scale sampling and scale-aware fusion strategies to enable the grounding model to efficiently accommodate the sparser and in-the-wild outdoor point cloud scenes.

\subsection{Comparative Experiments}
\textbf{Comparison with State-of-the-art Methods.}
We report in Tab. \ref{tab:sota_comp} the benchmark performance of state-of-the-art (SoTA) 3DVG methods on the \textbf{\emph{MoT3DVG}} dataset.
Similar to the traditional 3DVG paradigm \cite{scanrefer, 3dvg-transformer, bdetr, eda}, we evaluate the benchmark results from the \emph{Unique} subset, \emph{Multiple} subset, and \emph{Overall}.
In the \emph{Unique} subset, only one object whose category matches the referring expression is present in a single frame, whereas the \emph{Multiple} subset contains multiple semantically similar objects. The \emph{Overall} is the aggregated evaluation across the entire dataset.
As shown in Tab. \ref{tab:sota_comp}, our DynaVG surpasses both indoor and outdoor SoTA  3DVG methods after incorporating temporal cues.
Compared to 3EED \cite{3eed}, our DynaVG improves Acc@0.5 by 12.29\% and 2.40\% on the \emph{Multiple} and \emph{Overall} settings, respectively.
Note that in \emph{MoT3DVG}, most samples fall into the \emph{Multiple} subset, while only a minority is \emph{Unique}. Owing to this imbalance, the model is biased towards learning from \emph{Multiple} conditions.

\textbf{Comparison Across Various Ranges.}
To investigate the impact of object distance on grounding performance, we partitioned the dataset into three intervals according to the valid detection range (0-20 m, 20-40, 40-50 m).
As shown in Tab. \ref{tab:dist_compare}, grounding accuracy continues to decrease with distance. 
By incorporating temporal cues from point cloud sequences, DynaVG effectively mitigates performance degradation, particularly under \emph{Multiple} settings.
However, we observe a degradation in accuracy under the \emph{Unique} setting.
One possible reason is that temporal aggregation introduces slight temporal lags, smoothing the current-frame features with historical cues, which may affect accuracy when only one candidate exists.
Meanwhile, since the \emph{Unique} split is relatively small, its results are more sensitive to minor variations.
We provide more detailed analyzes of this in the appendix \ref{sec: failure_cases}.



\textbf{Comparison on 3EED Dataset.}
To further validate the effectiveness of our proposed DynaVG, we compare it against the outdoor SoTA 3DVG methods on the 3EED \emph{Vehicle} dataset \cite{3eed} in Tab. \ref{tab:3eed_compare}. 
For a  fair comparison, we adopt the 3EED protocol and evaluate at IoU thresholds of 0.25 and 0.5.
As shown in the table, equipping the model with dynamic information encoding enables DynaVG to consistently improve outdoor 3DVG performance, yielding gains of +1.65\% and +1.25\% on Acc@0.25 and Acc@0.5 over 3EED, respectively.
Furthermore, to enable an mIoU comparison, we re-implemented 3EED using its official open-source codebase, and our DynaVG attains 45.41\% mIoU, surpassing 3EED by 1.47\%.

\textbf{Class-wise Comparison.}
To investigate the influence of semantic category on grounding performance, we further evaluate the fine-grained categories \emph{vehicle-car} and \emph{vehicle-bus} within the coarse category \emph{Vehicle}.
As detailed in Tab. \ref{tab:class_wise_compare}, our DynaVG improves Acc@0.5 by 14.06\% and 1.72\% on \emph{vehicle-car} and \emph{vehicle-bus}, respectively.
Due to differences in data volume and instance scale, the model tends to learn patterns of \emph{vehicle-car} more readily. Nevertheless, we are still able to improve grounding accuracy to some extent through temporal cues.

\begin{wraptable}{r}{0.55\textwidth}
\centering
\vspace{-12.5pt}
\caption{Comparison with state-of-the-art 3DVG methods on different view. `S.Left', `S.Right' are the views of the side left and side right, respectively.}
\resizebox{0.55\textwidth}{!}{
\begin{tabular}{c|c|ccc}
\toprule
Method      & View                     & @1.0 & @0.5 & mIoU \\
\midrule \midrule
3EED \cite{3eed}       & \multirow{3}{*}{Front}   & 61.84        & 46.11        & 20.76     \\
DynaVG (Ours)     &                          & 65.25        & 51.06     & 21.63     \\
\emph{Improvement} &                          & +3.41        & +4.95        & +0.87     \\ 
\midrule
3EED \cite{3eed}        & \multirow{3}{*}{S.Left}  & 60.06        & 48.90        & 19.46     \\
DynaVG (Ours)       &                          & 63.33        & 50.25        & 20.02     \\
\emph{Improvement} &                          & +3.27        & +1.35        & +0.54     \\ 
\midrule
3EED \cite{3eed}        & \multirow{3}{*}{S.Right} & 66.75        & 53.91        & 24.48     \\
DynaVG (Ours)       &                          & 68.64        & 54.07        & 25.34     \\
\emph{Improvement} &                          & +1.89 & +0.16   & +0.86     \\ 
\midrule
3EED \cite{3eed}        & \multirow{3}{*}{Full} & 67.22        & 54.94        & 21.16     \\
DynaVG (Ours)       &                          & 69.66        & 57.34        & 22.22     \\
\emph{Improvement} &                          & +2.44        & +2.40        & +1.06     \\
\bottomrule
\end{tabular}
}
\vspace{-0.6em}
\label{tab:view_ablation}
\end{wraptable}

\textbf{Multi-view Comparison.}
In the preceding experiments, we adopted a full-field-of-view setting.
Inspired by the 3EED dataset \cite{3eed}, we investigate the influence of different viewpoints on the performance of the outdoor grounding model.
Following the camera viewpoints and extrinsic–intrinsic matrices defined in nuScenes \cite{nuscenes}, we partition the point cloud into multiple view frusta: \emph{front, front left, front right, back, back right, and back left}.
In Tab. \ref{tab:view_ablation}, we report grounding results for the viewpoints of front, side left (front left and back left), and side right (front right and back right). 
As shown in the table, the introduction of temporal cues effectively enriches the current-frame representation, alleviating accuracy limitations arising from viewpoint occlusion.

\subsection{Ablation Study}

\textbf{Impact of Short-Term Window Encoding (STWE).}
This strategy aims to capture local temporal dynamic cues by aggregating features only within locally adjacent frames in the same window.
To isolate its contribution, we disable the subsequent cross-window interactions.
As shown in the second row of Tab. \ref{tab:ablation_study}, after intra-window temporal encoding, the grounding model improves by 0.19\%, 0.26\%, and 0.25\% on Acc@1.0, Acc@0.5, and mIoU, respectively, over the baseline, confirming the efficacy of incorporating temporal motion cues.

\begin{wraptable}{r}{0.45\textwidth}
\centering
\vspace{-12.5pt}
\caption{Ablation study on components of the SLIDE module.}
\resizebox{0.45\textwidth}{!}{
\footnotesize
\begin{tabular}{cc|ccc}
\toprule
STWE & LTWS & @1.0 & @0.5 & mIoU \\ 
\midrule \midrule
     &      & 67.22  & 54.94  & 21.16 \\
\ding{51} &      & 67.41  & 55.20  & 21.41 \\
     & \ding{51} & 67.71 & 55.23 & 21.71 \\
\ding{51} & \ding{51} & 69.66 & 57.34 & 22.22 \\
\bottomrule
\end{tabular}
}
\label{tab:ablation_study}
\end{wraptable}

\textbf{Impact of Long-Term Window Shifting (LTWS).}
The primary goal of LTWS is to model global temporal cues and integrate them into the current-frame features.
In this ablation, we remove STWE and instead encode all historical frames in a single window to capture long-range temporal cues.
As shown in the third row of Tab.~\ref{tab:ablation_study}, this strategy also improves accuracy compared to the baseline.
However, the data comparison reveals that the accuracy gains from either the STWE or this long-range-only variant alone are quite limited.
In the STWE strategy, we model local motion cues between adjacent frames but lack global temporal context. Conversely, directly modeling global temporal information tends to overlook local motion details.
Therefore, we adopt a window shifting strategy to capture global temporal motion cues while preserving local details, effectively enhancing model performance while maintaining computational efficiency.

\section{Conclusion}
In this paper, we present \textbf{\emph{MoT3DVG}}, a large-scale benchmark for motion-aware outdoor 3D visual grounding.
Built upon nuScenes, \textbf{\emph{MoT3DVG}} provides over 985,000 objects and 144,568 motion-aware language descriptions for dynamic objects across 31,128 frames from 850 scenes, covering fine-grained motion behaviors together with object appearance and spatial relations.
In addition, we propose \textbf{DynaVG}, a dynamic-aware framework for outdoor 3DVG that leverages multi-frame temporal cues.
Specifically, subsequent to the conventional point cloud encoder, we introduce a SLIDE module that encodes both local and long-range temporal cues via short-term window encoding and long-term window shifting strategies. 
SLIDE then injects these temporal cues into the current-frame features via cross-attention, enabling alignment with motion-aware descriptions in the language prompts.
Based on \textbf{\emph{MoT3DVG}} dataset and DynaVG, we benchmark SoTA 3DVG methods under multiple settings, demonstrating the importance of incorporating temporal cues for motion-aware outdoor grounding. We hope that our work will facilitate advancements in outdoor 3DVG research.

\section{Acknowledgements}
This work was supported by the National Natural Science Foundation of China (No.42571514).

\bibliographystyle{unsrtnat}
\bibliography{reference}

\newpage
\appendix

\section{Supplement to \emph{\textbf{MoT3DVG}} Dataset}
In this section, we provide additional details on the \emph{\textbf{MoT3DVG}} dataset, including the preparatory work for motion-language annotation and dataset statistics.

\begin{figure*}[h]
    \centering
    \includegraphics[width=1.0\linewidth]{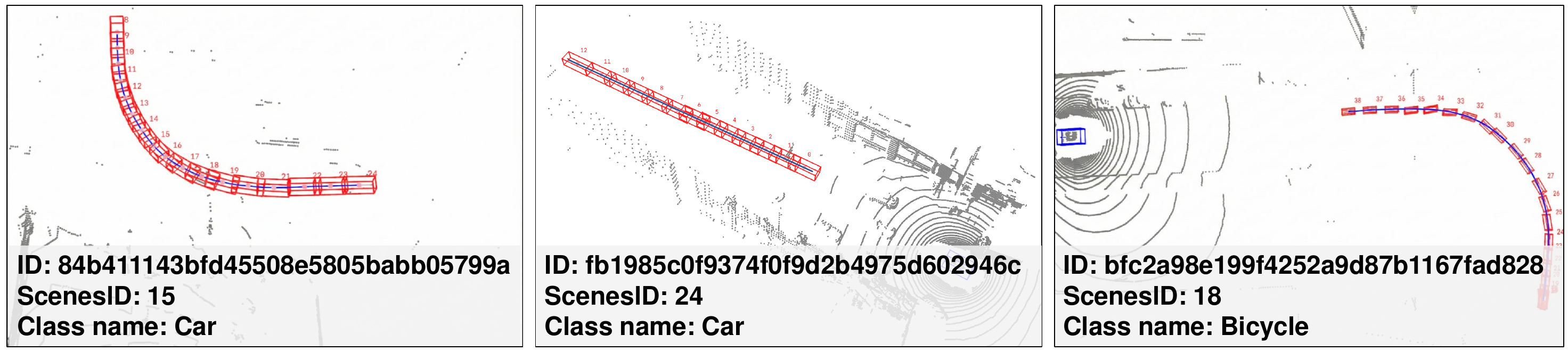}
    \caption{Visualization of instance trajectories. Each red box denotes a timestep. ID denotes the current instance token, sceneID represents the index of the current scene, and class name indicates the category of the current object.
}
    \label{fig:traj_vis}
\end{figure*}

\subsection{Annotation Preparation and Protocol}\label{sec: supp_annotation}
\textbf{Annotation preparation.} In this part, we provide additional details regarding the preparatory work conducted before motion-aware language annotation. 
For each nuScenes scene sequence of length $T$, we first parse the official object detection annotations \cite{nuscenes} to construct a state representation for each object instance $i$. Specifically, at every timestamp $t$, we define the object state as a tuple $\mathbf{s}_{t}^{(i)} = \{c^{(i)}, \mathbf{p}_{t}^{(i)}, \mathbf{v}_{t}^{(i)}\}$, where $c^{(i)}$ denotes the semantic category and instance ID, $\mathbf{p}_{t}^{(i)} \in \mathbb{R}^3$ represents the 3D centroid coordinates, and $\mathbf{v}_{t}^{(i)}$ is the velocity vector. We subsequently classify the motion status of each instance, filtering out those that remain static throughout the sequence. For the remaining dynamic instances, we unify their per-timestep positions into a canonical coordinate system anchored at the initial frame ($t=0$), forming a continuous trajectory set $\mathcal{T}_{traj}^{(i)} = \{ \mathbf{p}_{0}^{(i)}, \dots, \mathbf{p}_{T-1}^{(i)} \}$. To address temporal sparsity caused by occlusions or missing annotations, we employ linear interpolation to complete the trajectory. For a missing frame $t'$ bounded by the nearest valid timestamps $t_a$ and $t_b$ (where $t_a < t' < t_b$), the estimated position $\hat{\mathbf{p}}_{t'}^{(i)}$ is computed as:
\begin{equation}
\hat{\mathbf{p}}_{t'}^{(i)} = \mathbf{p}_{t_a}^{(i)} + \frac{t' - t_a}{t_b - t_a} \left( \mathbf{p}_{t_b}^{(i)} - \mathbf{p}_{t_a}^{(i)} \right).
\end{equation}

Upon completing the preparatory process illustrated in Fig. \ref{fig:traj_vis}, we can proceed to motion state annotation with enhanced efficiency.

\textbf{Annotation protocol.} For each object, annotators were provided with its BEV trajectory, per-timestep $x$--$y$ positions, heading, and velocity estimates. All annotation groups followed shared guidelines containing definitions and representative trajectory examples for straight motion, turning, lane changing, acceleration, and deceleration.
For directional labels, straight motion was assigned when the heading remained approximately stable, and the lateral displacement stayed within approximately one object length. Turning required sustained lateral displacement accompanied by a clear heading change, whereas lane changing referred to lateral displacement while the overall travel heading remained approximately stable. For speed labels, annotators jointly examined the nuScenes velocity estimates and observed inter-frame trajectory changes to identify acceleration and deceleration. They also marked the timestep at which each motion-state transition occurred.
After the initial annotation, labels were cross-checked between different annotation groups to examine consistency.

\subsection{Supplement to Data Statistic}\label{sec: supp_data_statistic}
\textbf{Motion components.}
When annotating motion-language descriptions, we categorize them into directional motion and velocity-based motion.
Directional motion denotes the movement direction of the object in 3D environments, such as "moving straight", "turning left or right", and "lane changing".
Velocity-based motion describes the speed variations of the object, such as "speeding up", "slowing down", and "stopping".
Tab. \ref{tab:supp_mot_component} summarizes the statistics of motion descriptions in the training and validation part of \emph{\textbf{MoT3DVG}} dataset.


\begin{table*}[t]
    \centering
    \begin{minipage}{0.48\textwidth}
        \centering
        \vspace{0pt}
        \caption{Statistics of motion descriptions in \emph{MoT3DVG}.}
        \label{tab:supp_mot_component}
        
        \small
        \setlength{\tabcolsep}{2mm}
        \begin{tabular}{c|c|c|c}
            \toprule
            & Split & Scenes & Expression \\ 
            \midrule\midrule
            Direction & \multirow{3}{*}{Training} & \multirow{3}{*}{700} & 106,477 \\
            Speed & & & 23,807 \\
            Total & & & 117,072 \\ 
            \midrule
            Direction & \multirow{3}{*}{Validation} & \multirow{3}{*}{150} & 24,521 \\
            Speed & & & 4,745 \\
            Total & & & 27,496 \\ 
            \midrule
            Summary & Overall & 850 & 144,568 \\ 
            \bottomrule
        \end{tabular}
    \end{minipage}
    \hfill
    \begin{minipage}{0.48\textwidth}
        \centering
        \vspace{0pt}
        \includegraphics[width=\linewidth]{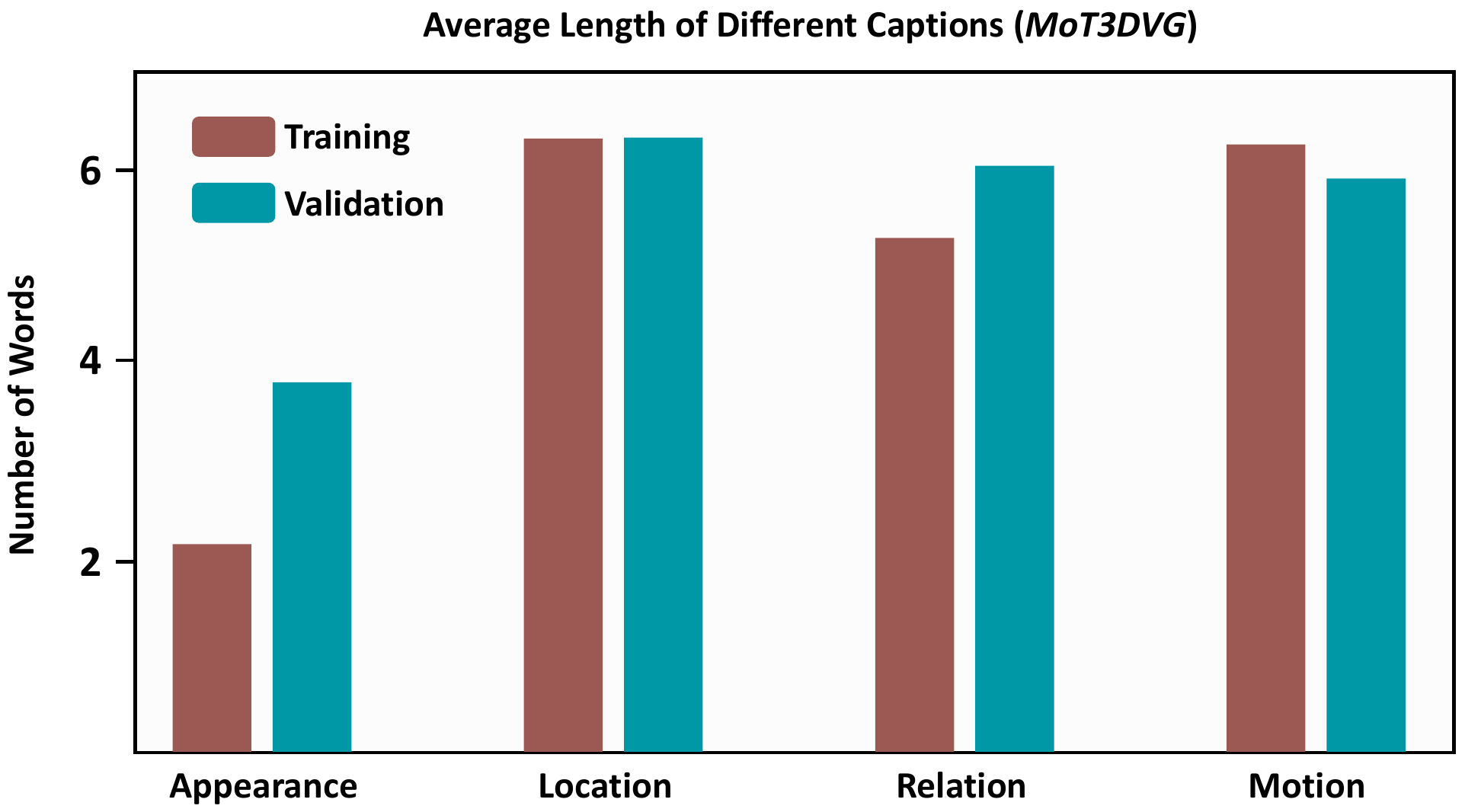}
        \captionof{figure}{Average length of different captions.}
        \label{fig:supp_average_words}
    \end{minipage}
\end{table*}

\begin{figure*}[t]
  \centering
   \includegraphics[width=0.8\linewidth]{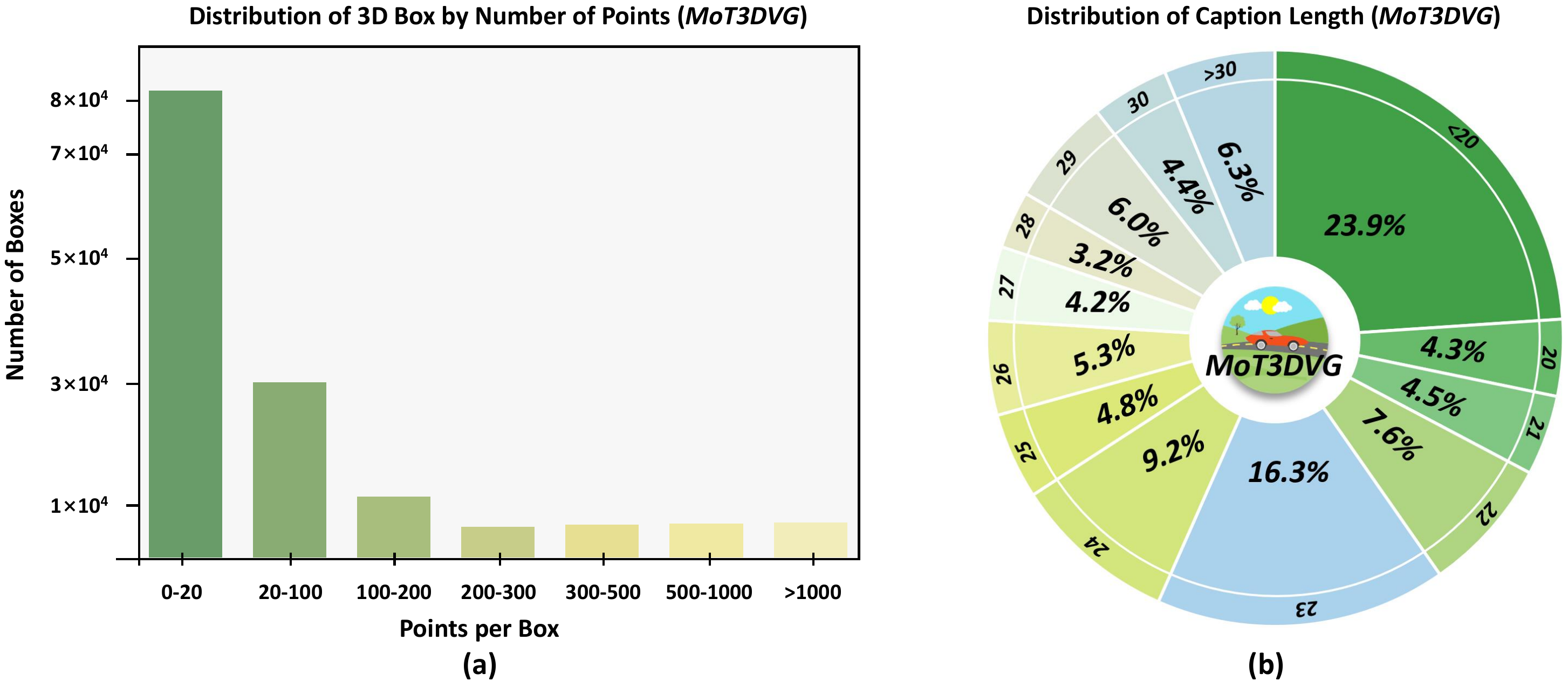}
   \caption{Dataset statistics of \emph{\textbf{MoT3DVG}}. (a) Distribution of boxes by number of points. (b) Distribution of caption length. The numbers in the outer ring denote the word lengths contained within the captions, while the numbers in the inner ring indicate the proportion of each length across the entire dataset.
   }
   \label{fig:supp_statistic_vis}
\end{figure*}

\textbf{Other Statistics.}
In addition to motion descriptions, the textual annotations in \emph{\textbf{MoT3DVG}} also encompass language descriptions of the object's appearance attributes, location attributes, and relationships with other objects, which are derived from the reorganization of TOD$^3$Cap \cite{tod3cap} captions. 
We visualize the distribution of average caption lengths across the training and validation splits in Fig. \ref{fig:supp_average_words}.
In Fig. \ref{fig:supp_statistic_vis} (a), we also analyze the distribution of point cloud counts within the 3D bounding boxes.
As the nuScenes dataset \cite{nuscenes} was collected using a 32-beam LiDAR sensor, a considerable proportion of 3D bounding boxes contain fewer than 20 point clouds.
To prevent the model from being affected during training, we filter out bounding boxes with fewer than 20 interior points.
Besides, in Tab. \ref{fig:supp_statistic_vis} (b), we analyze the distribution of word lengths in the language descriptions.

\subsection{Linguistic Diversity Analysis} 
\textbf{Language diversity.} MoT3DVG exhibits higher language diversity compared with existing datasets \cite{3eed, tod3cap}. We additionally supplement the motion-aware descriptions of objects, which have been overlooked by all prior datasets. We provide quantitative evaluation results based on Herdan’s C \cite{herdan_c} in Tab. \ref{tab:supp_diversity}.

\textbf{Template bias.}  Existing methods mainly adopt template-based paragrams to generate language descriptions \cite{3eed, tod3cap}, but they typically focus on describing static object attributes while ignoring dynamic information in outdoor scenes. Following \cite{tod3cap}, we design generation templates for motion descriptions. Specifically, we first decompose each component within our defined direction and speed categories into fine-grained motions, and then generate diverse sub-templates using an LLM for instantiation.
Although an LLM generates diverse sub-templates, the complete motion descriptions are still assembled from predefined directional and speed components. This compositional structure can introduce template bias and may produce less natural expressions than free-form human referring expressions.
Despite the diversity of the sub-templates, we acknowledge that the complete motion descriptions are composed using predefined templates and may therefore be less natural than free-form human referring expressions.

\begin{table*}[htbp]
\caption{
Linguistic diversity analysis.
}
\centering
    \small
    \begin{tabular}{c|c}
    \toprule
    Dataset               & Herdan's C $\uparrow$ \\
    \midrule \midrule
    3EED \cite{3eed}      & 0.9052 \\
    TOD3Cap \cite{tod3cap}& 0.9190 \\
    \textbf{MoT3DVG}      & \textbf{0.9401} \\
    \bottomrule
\end{tabular}
\label{tab:supp_diversity}
\end{table*}

\section{Loss Function}
During the training of 3DVG models, the key point lies in achieving semantic alignment between the input language prompt and the point cloud object.
To this end, the model first constructs the overall cost matrix by aggregating semantic classification costs $C_{cls}$, bounding box L1 distance costs $C_{l_1}$, and GIoU costs $C_{giou}$:
\begin{equation}
C = \lambda_{cls} C_{cls} + \lambda_{l_1} C_{l_1} + \lambda_{giou} C_{giou},
\label{eq:cost_matrix}
\end{equation}
where $\lambda_{cls}$, $\lambda_{l_1}$, and $\lambda_{giou}$ denote the balancing weights used to regulate different cost terms, and we set their values to 1, 0, and 2, similar to \cite{3eed}.
We then construct the Hungarian matching based on the cost matrix to match predicted object queries with ground-truth boxes. 
Based on the matched positive pairs, we achieve semantic alignment through the following contrastive learning losses:
\begin{equation}
\mathcal{L}_{p2t} = -\frac{1}{N_{pos}}\sum_{(i,j)\in \mathcal{I}_{pos}}\log\frac{\exp(f^i_{p}\cdot (f^j_t)^T/ \tau)}{\sum_{j'}\exp(f^i_{p}\cdot (f^{j'}_t)^T/ \tau)},
\end{equation}
\begin{equation}
\mathcal{L}_{t2p} = -\frac{1}{N_{pos}}\sum_{(i,j)\in \mathcal{I}_{pos}}\log\frac{\exp(f^i_{p}\cdot (f^j_t)^T/ \tau)}{\sum_{i'}\exp(f^{i'}_{p}\cdot (f^j_t)^T/ \tau)},
\end{equation}
where $\mathcal{I}_{pos}$ denotes the index set of positive matched point cloud-language pairs and $N_{pos}=|\mathcal{I}_{pos}|$.
Here $i$ indexes the point cloud feature $f_p$ in a positive pair, and $j$ indexes the matched language feature $f_t$. In the denominators, $j'$ ranges over all language features in the mini-batch (negatives for fixed $i$) and $i'$ ranges over all point cloud features in the mini-batch (negatives for fixed $j$). $\tau$ is the temperature factor (set to 0.07).
The model training is jointly supervised by classification loss, bounding box regression loss, GIoU loss, and contrastive learning loss.

\section{Additional Experiments}
\subsection{Comparison with Different Motion Descriptions}
To validate the necessity of incorporating temporal motion-aware language descriptions, we conduct additional comparative experiments by reorganizing the language prompts using the coarse motion information provided by TOD$^3$Cap \cite{tod3cap}.
We report the results in the Tab. \ref{tab:supp_mot_compare}, validating the effectiveness of fine-grained motion-aware language descriptions over coarse motion ones.
Meanwhile, we believe that providing fine-grained motion annotations along the temporal axis can facilitate the development of downstream outdoor 3DVG tasks, such as video grounding in outdoor 3D scenes, which we will further explore in our future work.

\begin{table*}[htbp]
\caption{
Comparison with different motion descriptions.
}
\centering
\small
\setlength{\tabcolsep}{3mm}
\begin{tabular}{c|ccc}
\toprule
                & Acc@1.0 & Acc@0.5 & mIoU \\ \midrule \midrule
Coarse motion   & 66.61        &   54.76      & 21.37     \\
Temporal motion & 69.66        & 57.34        & 22.22     \\ \bottomrule
\end{tabular}
\label{tab:supp_mot_compare}
\end{table*}

\subsection{Addational Ablation Study}
\textbf{Ablation study on language prompt composition.}
To further verify the impact of motion-aware descriptions on outdoor 3DVG models, we conduct ablation experiments on the constituent attributes of language prompts. As shown in Tab. \ref{tab:supp_ablation_app_motione}, removing appearance attributes (following 3EED \cite{3eed}, we only use point cloud as visual input, and the absence of RGB information introduces interference brought by color and texture attributes) does not lead to an obvious performance change of the grounding model, while removing motion attributes results in a significant performance drop. This indicates that the model relies more on motion information when grounding the object.

\begin{table*}[htbp]
\caption{
Ablation study on language prompt composition.
}
\centering
\small
\setlength{\tabcolsep}{3mm}
\begin{tabular}{c|ccc}
\toprule
    Model                & Acc@1.0 & Acc@0.5 & mIoU  \\
    \midrule \midrule
    Full                 & 69.66   & 57.34   & 22.22 \\
    without appearance   & 69.15   & 58.63   & 21.71 \\
    without motion       & 67.05   & 54.33   & 21.44 \\ \bottomrule
\end{tabular}
\label{tab:supp_ablation_app_motione}
\end{table*}

\textbf{Multimodal ablation study.}
Our primary evaluation follows a LiDAR-centric setting to study temporal motion cues in outdoor 3DVG. In this ablation, we additionally render RGB-derived image semantics into the point cloud features to assess whether DynaVG remains effective when appearance information is available. As shown in Tab. \ref{tab:supp_multimodal}, adding image semantics yields modest improvements over the LiDAR-only configuration. These results characterize the effect of incorporating appearance cues in the evaluated configuration.

\begin{table*}[htbp]
\caption{
Multimodal ablation study.
}
\centering
    \small
    \begin{tabular}{c|ccc}
    \toprule
    Modality & Acc@1.0 & Acc@0.5 & mIoU  \\
    \midrule \midrule
    L        & 69.66   & 57.34   & 22.22 \\
    L+I      & 70.07   & 58.20   & 22.67 \\
    \bottomrule
\end{tabular}
\label{tab:supp_multimodal}
\end{table*}

\textbf{Additional ablation study on SLIDE.}
To verify the effectiveness of the SLIDE module in exploiting temporal cues, we have incorporated multi-frame information into 3EED \cite{3eed} and conducted the ablation study. As shown in Tab. \ref{tab:supp_temporal}, simply introducing temporal cues is insufficient to achieve optimal performance. By short-term window encoding and long-term window shifting, SLIDE captures both local and global temporal cues, effectively enhancing current-frame features.

\begin{table*}[htbp]
\caption{
Additional ablation study of SLIDE.
}
\centering
    \small
    \begin{tabular}{c|ccc}
    \toprule
    Method                          & Acc@1.0 & Acc@0.5 & mIoU  \\
    \midrule \midrule
    3EED \cite{3eed}                & 67.22   & 54.94   & 21.16 \\
    Historical frame concatenate    & 67.29   & 56.04   & 21.66 \\
    \textbf{SLIDE (Ours)}            & 69.66   & 57.34   & 22.22 \\
    \bottomrule
\end{tabular}
\label{tab:supp_temporal}
\end{table*}

\textbf{Adaptive temporal gating.}
We add a lightweight gate that weights historical features using current-frame point density and distance statistics. As shown in Tab.~\ref{tab:supp_adaptive_gate}, gating improves Overall performance within 0--40\,m, but its effects are less consistent across metrics for Unique objects and at longer distances. These mixed results suggest that density and distance alone may not reliably determine when historical features are useful.
\begin{table*}[htbp]
\caption{
Effect of adaptive temporal gating across different distance ranges.
All metrics are reported in percentage (\%).
}
\centering
    \small
    \setlength{\tabcolsep}{2.5pt}
    \begin{tabular}{c|ccc|ccc|ccc}
    \toprule
    Method
    & \multicolumn{3}{c|}{Unique}
    & \multicolumn{3}{c|}{Multiple}
    & \multicolumn{3}{c}{Overall} \\
    & Acc@1.0 & Acc@0.5 & mIoU
    & Acc@1.0 & Acc@0.5 & mIoU
    & Acc@1.0 & Acc@0.5 & mIoU \\
    \midrule \midrule
    \multicolumn{10}{c}{0--20\,m} \\
    \midrule
    3EED \cite{3eed}
    & 57.08 & 43.38 & 17.88
    & 64.64 & 56.36 & 18.73
    & 73.23 & 64.55 & 21.46 \\
    DynaVG
    & 50.68 & 38.81 & 19.02
    & 75.95 & 67.43 & 22.02
    & 74.22 & 64.87 & 21.81 \\
    DynaVG+gate
    & 54.59 & 38.53 & 18.93
    & 76.89 & 68.72 & 22.37
    & 75.21 & 66.40 & 22.11 \\
    \midrule
    \multicolumn{10}{c}{20--40\,m} \\
    \midrule
    3EED \cite{3eed}
    & 43.79 & 18.30 & 14.45
    & 48.95 & 32.38 & 18.32
    & 57.02 & 37.24 & 21.27 \\
    DynaVG
    & 52.29 & 21.57 & 19.35
    & 59.71 & 43.71 & 22.25
    & 59.93 & 40.32 & 22.51 \\
    DynaVG+gate
    & 55.84 & 20.13 & 19.10
    & 62.00 & 44.67 & 23.33
    & 61.23 & 41.60 & 22.82 \\
    \midrule
    \multicolumn{10}{c}{40--50\,m} \\
    \midrule
    3EED \cite{3eed}
    & 10.00 & 6.67 & 5.40
    & 15.28 & 5.56 & 8.81
    & 12.12 & 6.06 & 11.41 \\
    DynaVG
    & 26.67 & 6.67 & 31.55
    & 23.61 & 5.56 & 15.23
    & 20.20 & 6.06 & 17.13 \\
    DynaVG+gate
    & 14.29 & 3.57 & 26.17
    & 18.57 & 8.57 & 11.68
    & 17.35 & 7.14 & 15.82 \\
    \bottomrule
    \end{tabular}
\label{tab:supp_adaptive_gate}
\end{table*}

\textbf{Auxiliary motion supervision.}
We attach a lightweight motion-prediction head to the SLIDE-enhanced features and jointly regress the referent's velocity vector with the grounding objective.
As shown in Tab.~\ref{tab:supp_motion_head2}, auxiliary motion supervision improves Acc@1.0 and Acc@0.5 by 0.33 and 1.18 percentage points, respectively, while mIoU increases by 0.08 percentage points.
The additional head introduces 0.02M parameters and 0.07\, MiB of memory usage.
These results suggest a modest complementary benefit of explicit velocity supervision in the evaluated setting.
\begin{table*}[htbp]
\caption{
Effect of auxiliary motion supervision.
Parameter counts are reported in millions and memory usage in MiB.
}
\centering
    \small
    \setlength{\tabcolsep}{4pt}
    \begin{tabular}{c|ccc|cc}
    \toprule
    Method
    & Acc@1.0 & Acc@0.5 & mIoU
    & Parameters (M) & Memory (MiB) \\
    \midrule \midrule
    DynaVG
    & 69.66 & 57.34 & 22.22
    & 146.05 & 1012.42 \\
    DynaVG+motion head
    & 69.99 & 58.52 & 22.30
    & 146.07 & 1012.49 \\
    \bottomrule
    \end{tabular}
\label{tab:supp_motion_head2}
\end{table*}

\subsection{Qualitative Visualization}
\begin{figure}[h]
  \centering
   \includegraphics[width=0.75\linewidth]{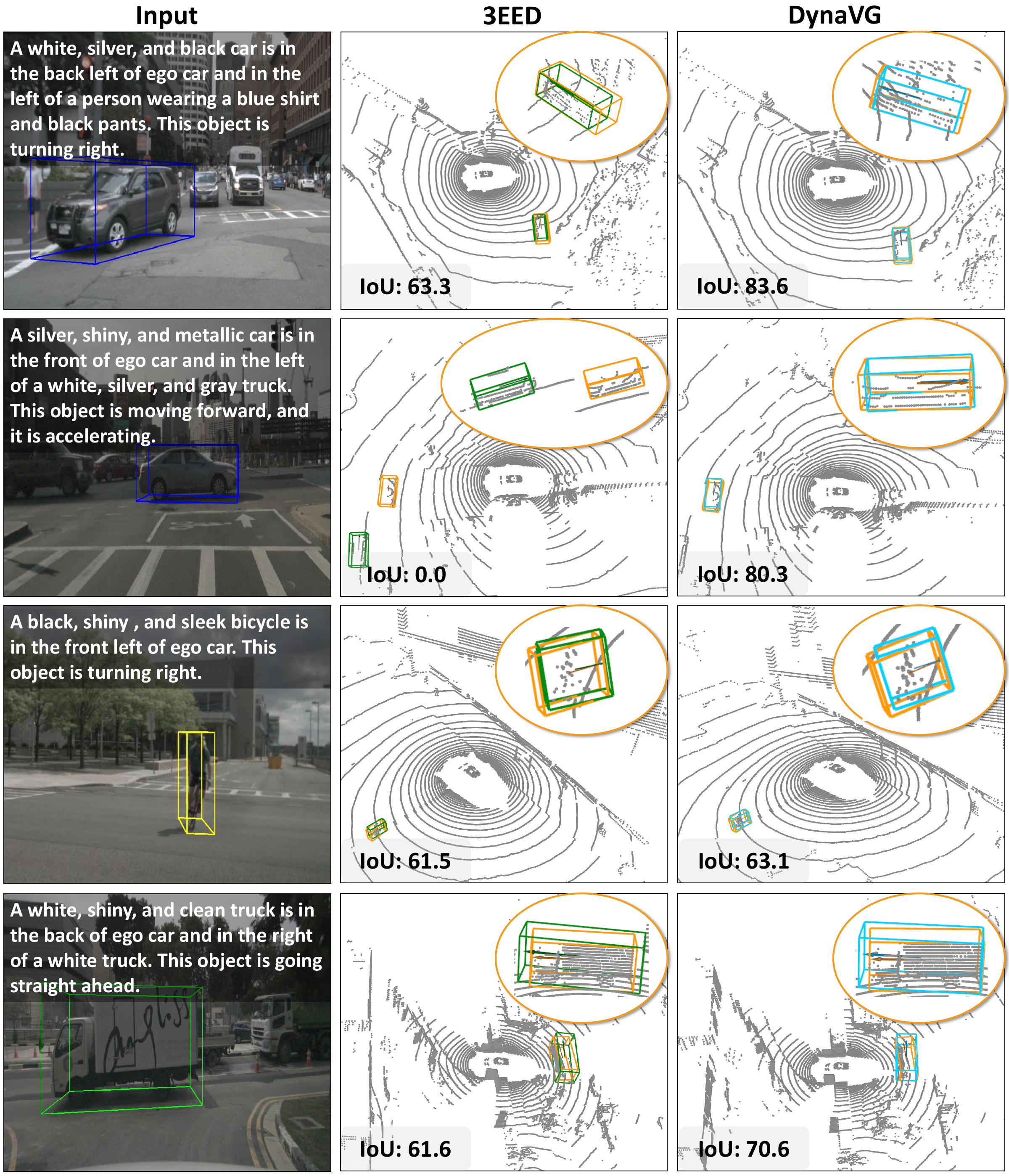}
   \caption{Qualitative visualization comparison with state-of-the-art outdoor 3DVG methods. The data are displayed with RGB frames and LiDAR point clouds, along with the corresponding language description.
   The ground truth boxes are given in orange. We use green and blue for the prediction boxes of 3EED and DynaVG, respectively.
}
   \label{fig:supp_vis}
\end{figure}

We present qualitative comparisons between our method and the state-of-the-art outdoor 3DVG method 3EED \cite{3eed} in Fig. \ref{fig:supp_vis}.
For clear differentiation, we use orange to denote the ground-truth boxes, while green and blue represent the predicted bounding boxes from 3EED and our DynaVG, respectively, based on the input language prompts.
As illustrated in the figure, the outdoor 3DVG model achieves more accurate localization of language-referred objects after incorporating temporal cues.

\subsection{Failure Cases}\label{sec: failure_cases}
\begin{figure}[h]
  \centering
   \includegraphics[width=0.65\linewidth]{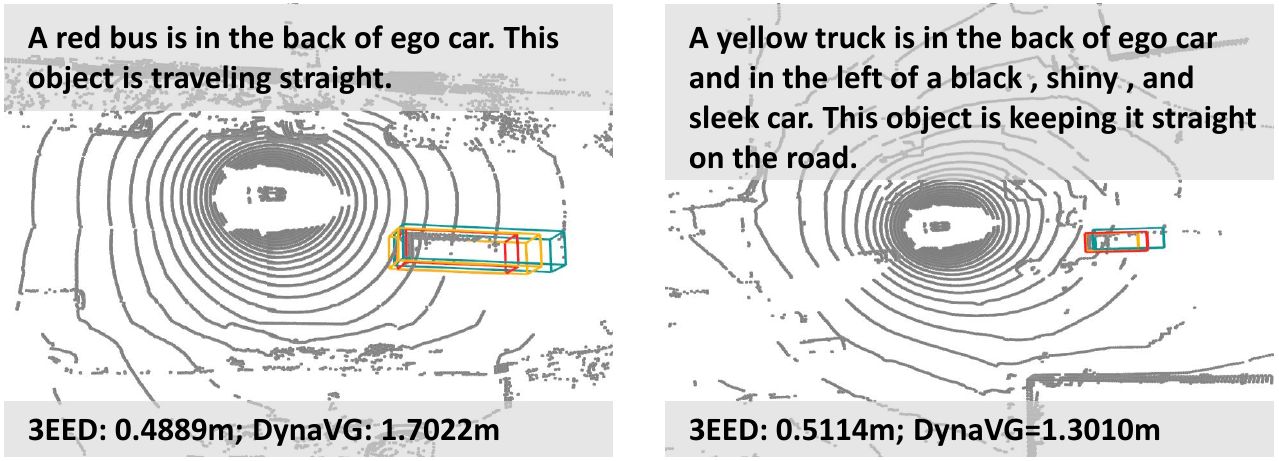}
   \caption{Visualization of failure cases. The ground truth boxes are given in red. We use orange and darkcyan for 3EED and DynaVG prediction boxes, respectively.
}
   \label{fig:supp_failure_cases}
\end{figure}

In this section, we analyze the failure causes of \emph{Unique} split in Tab. \ref{tab:dist_compare} from multiple perspectives.
In the \emph{Unique} split, semantic distractors are absent and near-field point clouds (0–20m) provide complete geometry. Static features alone can thus enable precise grounding. 
As shown in Fig. \ref{fig:supp_failure_cases}, additional temporal information may introduce spatial shifts and interfere with precise static features, which further leads to performance degradation.
In addition, we count the number of referent objects in the near-field \emph{Unique} split. 
As shown in Tab. \ref{tab:failure_case_count}, the referent object distribution in the \emph{Unique} split is relatively sparse, and minor prediction errors can inflate the overall evaluation metrics.

\begin{table*}[htbp]
\centering
\caption{Number count of referent objects in the near-field \emph{Unique} split.}
\label{tab:failure_case_count}
\small
\begin{tabular}{c|c|cc}
\toprule
\multirow{2}{*}{} & \multirow{2}{*}{Total num.} & \multicolumn{2}{c}{Positive num.} \\
                  &                       & 0.5         & 1.0         \\ \midrule \midrule
3EED \cite{3eed}             & \multirow{2}{*}{218}  & 101         & 128         \\
DynaVG            &                       &  87           &  111           \\
\bottomrule
\end{tabular}
\end{table*}

\section{Limitations and Perspectives}
\textbf{Limitations.} 
In this work, we verify the necessity of motion-aware language description for outdoor 3DVG tasks through extensive experiments.
However, we only design a concise temporal fusion module to implicitly align motion-aware language descriptions with visual temporal cues at the feature level, while lacking an explicit language-guided learning strategy for fine-grained motion description learning.
In addition, our work mainly focuses on single-vehicle platforms, and the collaborative utilization of motion information in multi-platform perception scenarios remains unexplored.
And, comprehensive multimodal fusion and dedicated modeling of extremely sparse long-range objects are not extensively explored in our LiDAR-centric setting.

\textbf{Future work.}
Based on the MoT3DVG dataset, we believe that further exploration of fine-grained motion understanding in outdoor scenarios is a promising research direction. It can provide crucial perceptual priors for outdoor autonomous driving systems and further advance the development of outdoor scene understanding as well as downstream tasks such as VLA and VLN.




\clearpage
\section*{NeurIPS Paper Checklist}

The checklist is designed to encourage best practices for responsible machine learning research, addressing issues of reproducibility, transparency, research ethics, and societal impact. Do not remove the checklist: {\bf The papers not including the checklist will be desk rejected.} The checklist should follow the references and follow the (optional) supplemental material.  The checklist does NOT count towards the page
limit. 

Please read the checklist guidelines carefully for information on how to answer these questions. For each question in the checklist:
\begin{itemize}
    \item You should answer \answerYes{}, \answerNo{}, or \answerNA{}.
    \item \answerNA{} means either that the question is Not Applicable for that particular paper or the relevant information is Not Available.
    \item Please provide a short (1--2 sentence) justification right after your answer (even for \answerNA). 
\end{itemize}

{\bf The checklist answers are an integral part of your paper submission.} They are visible to the reviewers, area chairs, senior area chairs, and ethics reviewers. You will also be asked to include it (after eventual revisions) with the final version of your paper, and its final version will be published with the paper.

The reviewers of your paper will be asked to use the checklist as one of the factors in their evaluation. While \answerYes{} is generally preferable to \answerNo{}, it is perfectly acceptable to answer \answerNo{} provided a proper justification is given (e.g., error bars are not reported because it would be too computationally expensive'' or ``we were unable to find the license for the dataset we used''). In general, answering \answerNo{} or \answerNA{} is not grounds for rejection. While the questions are phrased in a binary way, we acknowledge that the true answer is often more nuanced, so please just use your best judgment and write a justification to elaborate. All supporting evidence can appear either in the main paper or the supplemental material, provided in appendix. If you answer \answerYes{} to a question, in the justification please point to the section(s) where related material for the question can be found.

IMPORTANT, please:
\begin{itemize}
    \item {\bf Delete this instruction block, but keep the section heading ``NeurIPS Paper Checklist"},
    \item  {\bf Keep the checklist subsection headings, questions/answers and guidelines below.}
    \item {\bf Do not modify the questions and only use the provided macros for your answers}.
\end{itemize}


\begin{enumerate}

\item {\bf Claims}
    \item[] Question: Do the main claims made in the abstract and introduction accurately reflect the paper's contributions and scope?
    \item[] Answer: \answerYes{}{} 
    \item[] Justification: The primary focus in both the abstract and the introduction revolves around the description of the key contributions.
    \item[] Guidelines:
    \begin{itemize}
        \item The answer \answerNA{} means that the abstract and introduction do not include the claims made in the paper.
        \item The abstract and/or introduction should clearly state the claims made, including the contributions made in the paper and important assumptions and limitations. A \answerNo{} or \answerNA{} answer to this question will not be perceived well by the reviewers. 
        \item The claims made should match theoretical and experimental results, and reflect how much the results can be expected to generalize to other settings. 
        \item It is fine to include aspirational goals as motivation as long as it is clear that these goals are not attained by the paper. 
    \end{itemize}

\item {\bf Limitations}
    \item[] Question: Does the paper discuss the limitations of the work performed by the authors?
    \item[] Answer: \answerYes{} 
    \item[] Justification: We discuss the limitations of this work in the Limitations section of the appendix.
    \item[] Guidelines:
    \begin{itemize}
        \item The answer \answerNA{} means that the paper has no limitation while the answer \answerNo{} means that the paper has limitations, but those are not discussed in the paper. 
        \item The authors are encouraged to create a separate ``Limitations'' section in their paper.
        \item The paper should point out any strong assumptions and how robust the results are to violations of these assumptions (e.g., independence assumptions, noiseless settings, model well-specification, asymptotic approximations only holding locally). The authors should reflect on how these assumptions might be violated in practice and what the implications would be.
        \item The authors should reflect on the scope of the claims made, e.g., if the approach was only tested on a few datasets or with a few runs. In general, empirical results often depend on implicit assumptions, which should be articulated.
        \item The authors should reflect on the factors that influence the performance of the approach. For example, a facial recognition algorithm may perform poorly when image resolution is low or images are taken in low lighting. Or a speech-to-text system might not be used reliably to provide closed captions for online lectures because it fails to handle technical jargon.
        \item The authors should discuss the computational efficiency of the proposed algorithms and how they scale with dataset size.
        \item If applicable, the authors should discuss possible limitations of their approach to address problems of privacy and fairness.
        \item While the authors might fear that complete honesty about limitations might be used by reviewers as grounds for rejection, a worse outcome might be that reviewers discover limitations that aren't acknowledged in the paper. The authors should use their best judgment and recognize that individual actions in favor of transparency play an important role in developing norms that preserve the integrity of the community. Reviewers will be specifically instructed to not penalize honesty concerning limitations.
    \end{itemize}

\item {\bf Theory assumptions and proofs}
    \item[] Question: For each theoretical result, does the paper provide the full set of assumptions and a complete (and correct) proof?
    \item[] Answer: \answerNA{} 
    \item[] Justification: Our work incorporates only experimental results.
    \item[] Guidelines:
    \begin{itemize}
        \item The answer \answerNA{} means that the paper does not include theoretical results. 
        \item All the theorems, formulas, and proofs in the paper should be numbered and cross-referenced.
        \item All assumptions should be clearly stated or referenced in the statement of any theorems.
        \item The proofs can either appear in the main paper or the supplemental material, but if they appear in the supplemental material, the authors are encouraged to provide a short proof sketch to provide intuition. 
        \item Inversely, any informal proof provided in the core of the paper should be complemented by formal proofs provided in appendix or supplemental material.
        \item Theorems and Lemmas that the proof relies upon should be properly referenced. 
    \end{itemize}

    \item {\bf Experimental result reproducibility}
    \item[] Question: Does the paper fully disclose all the information needed to reproduce the main experimental results of the paper to the extent that it affects the main claims and/or conclusions of the paper (regardless of whether the code and data are provided or not)?
    \item[] Answer: \answerYes{} 
    \item[] Justification: We provide the specific settings of all hyper-parameters that appear in the main text, along with their corresponding experimental results.
    \item[] Guidelines:
    \begin{itemize}
        \item The answer \answerNA{} means that the paper does not include experiments.
        \item If the paper includes experiments, a \answerNo{} answer to this question will not be perceived well by the reviewers: Making the paper reproducible is important, regardless of whether the code and data are provided or not.
        \item If the contribution is a dataset and\slash or model, the authors should describe the steps taken to make their results reproducible or verifiable. 
        \item Depending on the contribution, reproducibility can be accomplished in various ways. For example, if the contribution is a novel architecture, describing the architecture fully might suffice, or if the contribution is a specific model and empirical evaluation, it may be necessary to either make it possible for others to replicate the model with the same dataset, or provide access to the model. In general. releasing code and data is often one good way to accomplish this, but reproducibility can also be provided via detailed instructions for how to replicate the results, access to a hosted model (e.g., in the case of a large language model), releasing of a model checkpoint, or other means that are appropriate to the research performed.
        \item While NeurIPS does not require releasing code, the conference does require all submissions to provide some reasonable avenue for reproducibility, which may depend on the nature of the contribution. For example
        \begin{enumerate}
            \item If the contribution is primarily a new algorithm, the paper should make it clear how to reproduce that algorithm.
            \item If the contribution is primarily a new model architecture, the paper should describe the architecture clearly and fully.
            \item If the contribution is a new model (e.g., a large language model), then there should either be a way to access this model for reproducing the results or a way to reproduce the model (e.g., with an open-source dataset or instructions for how to construct the dataset).
            \item We recognize that reproducibility may be tricky in some cases, in which case authors are welcome to describe the particular way they provide for reproducibility. In the case of closed-source models, it may be that access to the model is limited in some way (e.g., to registered users), but it should be possible for other researchers to have some path to reproducing or verifying the results.
        \end{enumerate}
    \end{itemize}

\item {\bf Open access to data and code}
    \item[] Question: Does the paper provide open access to the data and code, with sufficient instructions to faithfully reproduce the main experimental results, as described in supplemental material?
    \item[] Answer: \answerYes{} 
    \item[] Justification: We will open-source the dataset and code for this paper upon acceptance and provide guidance documents for reproducing the results.
    \item[] Guidelines:
    \begin{itemize}
        \item The answer \answerNA{} means that paper does not include experiments requiring code.
        \item Please see the NeurIPS code and data submission guidelines (\url{https://neurips.cc/public/guides/CodeSubmissionPolicy}) for more details.
        \item While we encourage the release of code and data, we understand that this might not be possible, so \answerNo{} is an acceptable answer. Papers cannot be rejected simply for not including code, unless this is central to the contribution (e.g., for a new open-source benchmark).
        \item The instructions should contain the exact command and environment needed to run to reproduce the results. See the NeurIPS code and data submission guidelines (\url{https://neurips.cc/public/guides/CodeSubmissionPolicy}) for more details.
        \item The authors should provide instructions on data access and preparation, including how to access the raw data, preprocessed data, intermediate data, and generated data, etc.
        \item The authors should provide scripts to reproduce all experimental results for the new proposed method and baselines. If only a subset of experiments are reproducible, they should state which ones are omitted from the script and why.
        \item At submission time, to preserve anonymity, the authors should release anonymized versions (if applicable).
        \item Providing as much information as possible in supplemental material (appended to the paper) is recommended, but including URLs to data and code is permitted.
    \end{itemize}

\item {\bf Experimental setting/details}
    \item[] Question: Does the paper specify all the training and test details (e.g., data splits, hyperparameters, how they were chosen, type of optimizer) necessary to understand the results?
    \item[] Answer: \answerYes{} 
    \item[] Justification: In the paper, we have clarified that we have adopted a consistent training and testing methodology, referring to mainstream approaches.
    \item[] Guidelines:
    \begin{itemize}
        \item The answer \answerNA{} means that the paper does not include experiments.
        \item The experimental setting should be presented in the core of the paper to a level of detail that is necessary to appreciate the results and make sense of them.
        \item The full details can be provided either with the code, in appendix, or as supplemental material.
    \end{itemize}

\item {\bf Experiment statistical significance}
    \item[] Question: Does the paper report error bars suitably and correctly defined or other appropriate information about the statistical significance of the experiments?
    \item[] Answer: \answerNo{} 
    \item[] Justification: The main experiments presented in the paper do not include error bars or statistical significance testing.
    \item[] Guidelines:
    \begin{itemize}
        \item The answer \answerNA{} means that the paper does not include experiments.
        \item The authors should answer \answerYes{} if the results are accompanied by error bars, confidence intervals, or statistical significance tests, at least for the experiments that support the main claims of the paper.
        \item The factors of variability that the error bars are capturing should be clearly stated (for example, train/test split, initialization, random drawing of some parameter, or overall run with given experimental conditions).
        \item The method for calculating the error bars should be explained (closed form formula, call to a library function, bootstrap, etc.)
        \item The assumptions made should be given (e.g., Normally distributed errors).
        \item It should be clear whether the error bar is the standard deviation or the standard error of the mean.
        \item It is OK to report 1-sigma error bars, but one should state it. The authors should preferably report a 2-sigma error bar than state that they have a 96\% CI, if the hypothesis of Normality of errors is not verified.
        \item For asymmetric distributions, the authors should be careful not to show in tables or figures symmetric error bars that would yield results that are out of range (e.g., negative error rates).
        \item If error bars are reported in tables or plots, the authors should explain in the text how they were calculated and reference the corresponding figures or tables in the text.
    \end{itemize}

\item {\bf Experiments compute resources}
    \item[] Question: For each experiment, does the paper provide sufficient information on the computer resources (type of compute workers, memory, time of execution) needed to reproduce the experiments?
    \item[] Answer: \answerYes{} 
    \item[] Justification: In the ‘Implementation Details' paragraph, we have specified the required computational resources.
    \item[] Guidelines:
    \begin{itemize}
        \item The answer \answerNA{} means that the paper does not include experiments.
        \item The paper should indicate the type of compute workers CPU or GPU, internal cluster, or cloud provider, including relevant memory and storage.
        \item The paper should provide the amount of compute required for each of the individual experimental runs as well as estimate the total compute. 
        \item The paper should disclose whether the full research project required more compute than the experiments reported in the paper (e.g., preliminary or failed experiments that didn't make it into the paper). 
    \end{itemize}
    
\item {\bf Code of ethics}
    \item[] Question: Does the research conducted in the paper conform, in every respect, with the NeurIPS Code of Ethics \url{https://neurips.cc/public/EthicsGuidelines}?
    \item[] Answer: \answerYes{} 
    \item[] Justification: The research conducted in the paper conforms, in every respect, to the NeurIPS Code of Ethics.
    \item[] Guidelines:
    \begin{itemize}
        \item The answer \answerNA{} means that the authors have not reviewed the NeurIPS Code of Ethics.
        \item If the authors answer \answerNo, they should explain the special circumstances that require a deviation from the Code of Ethics.
        \item The authors should make sure to preserve anonymity (e.g., if there is a special consideration due to laws or regulations in their jurisdiction).
    \end{itemize}

\item {\bf Broader impacts}
    \item[] Question: Does the paper discuss both potential positive societal impacts and negative societal impacts of the work performed?
    \item[] Answer: \answerYes{} 
    \item[] Justification: In the introduction, we have outlined the positive impact of our work.
    \item[] Guidelines:
    \begin{itemize}
        \item The answer \answerNA{} means that there is no societal impact of the work performed.
        \item If the authors answer \answerNA{} or \answerNo, they should explain why their work has no societal impact or why the paper does not address societal impact.
        \item Examples of negative societal impacts include potential malicious or unintended uses (e.g., disinformation, generating fake profiles, surveillance), fairness considerations (e.g., deployment of technologies that could make decisions that unfairly impact specific groups), privacy considerations, and security considerations.
        \item The conference expects that many papers will be foundational research and not tied to particular applications, let alone deployments. However, if there is a direct path to any negative applications, the authors should point it out. For example, it is legitimate to point out that an improvement in the quality of generative models could be used to generate Deepfakes for disinformation. On the other hand, it is not needed to point out that a generic algorithm for optimizing neural networks could enable people to train models that generate Deepfakes faster.
        \item The authors should consider possible harms that could arise when the technology is being used as intended and functioning correctly, harms that could arise when the technology is being used as intended but gives incorrect results, and harms following from (intentional or unintentional) misuse of the technology.
        \item If there are negative societal impacts, the authors could also discuss possible mitigation strategies (e.g., gated release of models, providing defenses in addition to attacks, mechanisms for monitoring misuse, mechanisms to monitor how a system learns from feedback over time, improving the efficiency and accessibility of ML).
    \end{itemize}
    
\item {\bf Safeguards}
    \item[] Question: Does the paper describe safeguards that have been put in place for responsible release of data or models that have a high risk for misuse (e.g., pre-trained language models, image generators, or scraped datasets)?
    \item[] Answer: \answerNA{} 
    \item[] Justification: The submitted paper poses no such risks.
    \item[] Guidelines:
    \begin{itemize}
        \item The answer \answerNA{} means that the paper poses no such risks.
        \item Released models that have a high risk for misuse or dual-use should be released with necessary safeguards to allow for controlled use of the model, for example by requiring that users adhere to usage guidelines or restrictions to access the model or implementing safety filters. 
        \item Datasets that have been scraped from the Internet could pose safety risks. The authors should describe how they avoided releasing unsafe images.
        \item We recognize that providing effective safeguards is challenging, and many papers do not require this, but we encourage authors to take this into account and make a best faith effort.
    \end{itemize}

\item {\bf Licenses for existing assets}
    \item[] Question: Are the creators or original owners of assets (e.g., code, data, models), used in the paper, properly credited and are the license and terms of use explicitly mentioned and properly respected?
    \item[] Answer: \answerYes{} 
    \item[] Justification: This paper utilizes open-source datasets, all of which have been properly licensed.
    \item[] Guidelines:
    \begin{itemize}
        \item The answer \answerNA{} means that the paper does not use existing assets.
        \item The authors should cite the original paper that produced the code package or dataset.
        \item The authors should state which version of the asset is used and, if possible, include a URL.
        \item The name of the license (e.g., CC-BY 4.0) should be included for each asset.
        \item For scraped data from a particular source (e.g., website), the copyright and terms of service of that source should be provided.
        \item If assets are released, the license, copyright information, and terms of use in the package should be provided. For popular datasets, \url{paperswithcode.com/datasets} has curated licenses for some datasets. Their licensing guide can help determine the license of a dataset.
        \item For existing datasets that are re-packaged, both the original license and the license of the derived asset (if it has changed) should be provided.
        \item If this information is not available online, the authors are encouraged to reach out to the asset's creators.
    \end{itemize}

\item {\bf New assets}
    \item[] Question: Are new assets introduced in the paper well documented and is the documentation provided alongside the assets?
    \item[] Answer: \answerYes{} 
    \item[] Justification: The discussions on new assets have been included in the appendix.
    \item[] Guidelines:
    \begin{itemize}
        \item The answer \answerNA{} means that the paper does not release new assets.
        \item Researchers should communicate the details of the dataset\slash code\slash model as part of their submissions via structured templates. This includes details about training, license, limitations, etc. 
        \item The paper should discuss whether and how consent was obtained from people whose asset is used.
        \item At submission time, remember to anonymize your assets (if applicable). You can either create an anonymized URL or include an anonymized zip file.
    \end{itemize}

\item {\bf Crowdsourcing and research with human subjects}
    \item[] Question: For crowdsourcing experiments and research with human subjects, does the paper include the full text of instructions given to participants and screenshots, if applicable, as well as details about compensation (if any)? 
    \item[] Answer: \answerNA{} 
    \item[] Justification: The motion annotations were completed internally by research collaborators and did not involve crowdsourcing or research with human subjects.
    \item[] Guidelines:
    \begin{itemize}
        \item The answer \answerNA{} means that the paper does not involve crowdsourcing nor research with human subjects.
        \item Including this information in the supplemental material is fine, but if the main contribution of the paper involves human subjects, then as much detail as possible should be included in the main paper. 
        \item According to the NeurIPS Code of Ethics, workers involved in data collection, curation, or other labor should be paid at least the minimum wage in the country of the data collector. 
    \end{itemize}

\item {\bf Institutional review board (IRB) approvals or equivalent for research with human subjects}
    \item[] Question: Does the paper describe potential risks incurred by study participants, whether such risks were disclosed to the subjects, and whether Institutional Review Board (IRB) approvals (or an equivalent approval/review based on the requirements of your country or institution) were obtained?
    \item[] Answer: \answerNA{} 
    \item[] Justification: The motion annotations were completed internally by research collaborators and did not involve crowdsourcing or research with human subjects.
    \item[] Guidelines:
    \begin{itemize}
        \item The answer \answerNA{} means that the paper does not involve crowdsourcing nor research with human subjects.
        \item Depending on the country in which research is conducted, IRB approval (or equivalent) may be required for any human subjects research. If you obtained IRB approval, you should clearly state this in the paper. 
        \item We recognize that the procedures for this may vary significantly between institutions and locations, and we expect authors to adhere to the NeurIPS Code of Ethics and the guidelines for their institution. 
        \item For initial submissions, do not include any information that would break anonymity (if applicable), such as the institution conducting the review.
    \end{itemize}

\item {\bf Declaration of LLM usage}
    \item[] Question: Does the paper describe the usage of LLMs if it is an important, original, or non-standard component of the core methods in this research? Note that if the LLM is used only for writing, editing, or formatting purposes and does \emph{not} impact the core methodology, scientific rigor, or originality of the research, declaration is not required.
    \item[] Answer: \answerNA{} 
    \item[] Justification: The core method development in this research does not involve LLMs as any important, original, or non-standard components.
    \item[] Guidelines:
    \begin{itemize}
        \item The answer \answerNA{} means that the core method development in this research does not involve LLMs as any important, original, or non-standard components.
        \item Please refer to our LLM policy in the NeurIPS handbook for what should or should not be described.
    \end{itemize}

\end{enumerate}

\end{document}